\documentclass{article} 
\usepackage{iclr2026_conference,times}

\usepackage{amsmath,amsfonts,bm}

\def\eqref#1{equation~\ref{#1}}

\def\1{\bm{1}}

\DeclareMathAlphabet{\mathsfit}{\encodingdefault}{\sfdefault}{m}{sl}
\SetMathAlphabet{\mathsfit}{bold}{\encodingdefault}{\sfdefault}{bx}{n}

\usepackage{wrapfig,lipsum}
\usepackage{hyperref}
\usepackage{url}

\usepackage{graphicx}
\usepackage{booktabs}   
\usepackage{multirow}   
\usepackage{array}  
\usepackage{adjustbox}   
\usepackage{multirow}
\usepackage{graphicx}
\usepackage{booktabs}    
\usepackage{tabularx}    
\usepackage{array}       
\usepackage{geometry}
\usepackage{algorithm}
\usepackage{hyperref}
\usepackage{url}
\usepackage[utf8]{inputenc}
\usepackage[T1]{fontenc}    
\usepackage{booktabs}       
\usepackage{amsfonts}       
\usepackage{nicefrac}       
\usepackage{microtype}      
\usepackage{bbm}
\usepackage{tabularx}
\usepackage{multirow}
\usepackage{colortbl}
\usepackage{wrapfig,lipsum}
\usepackage{makecell}
\usepackage{amsmath}
\usepackage{subfigure}
\usepackage[capitalize]{cleveref} 
\usepackage[T1]{fontenc}
\usepackage[utf8]{inputenc}
\usepackage{microtype}
\usepackage{inconsolata}
\usepackage{graphicx}
\usepackage{xcolor}
\definecolor{pygreen}{rgb}{0.0, 0.5, 0.0}
\usepackage[most]{tcolorbox}
\usepackage{pifont}
\usepackage{enumitem}

\definecolor{g}{gray}{0.75}
\title{The Quest for Efficient Reasoning: A Data-Centric Benchmark to CoT Distillation
}

\usepackage[framemethod=TikZ]{mdframed}
\usepackage[most]{tcolorbox}
\newtcolorbox{takeawaybox}{enhanced,
  colback=pink!20!gray!30!white, 
  colframe=pink!75!gray, 
}
\usepackage{caption}
\usepackage[table]{xcolor} 

\definecolor{ZSBaseline}{HTML}{ff8d13}
\definecolor{KDBaseline}{HTML}{bd00ff}
\definecolor{DABaseline}{HTML}{2782ed}
\definecolor{OurColor}{HTML}{36aa70}
\definecolor{UserExampleBg}{HTML}{ffffff}
\definecolor{UserExampleTitle}{HTML}{545f7f} 

\author{\textbf{Ruichen Zhang}\thanks{\ \ Equal contribution.}~~$^{1}$, \textbf{Rana Muhammad Shahroz Khan$^{*1}$}, \textbf{Zhen Tan$^{2}$},  \textbf{Dawei Li$^{2}$}, \textbf{Song Wang$^{3}$},\\
\textbf{Tianlong Chen}$^{1}$  \\
$^{1}$University of North Carolina at Chapel Hill,$^{2}$Arizona State University,\\
$^{3}$University of Virginia
}

\iclrfinalcopy 
\makeatletter
\def\ps@headings{
  \def\@oddhead{}\def\@evenhead{}
  \def\@oddfoot{\hfil\thepage\hfil}
  \def\@evenfoot{\hfil\thepage\hfil}
}
\def\ps@myheadings{
  \def\@oddhead{}\def\@evenhead{}
  \def\@oddfoot{\hfil\thepage\hfil}
  \def\@evenfoot{\hfil\thepage\hfil}
}
\makeatother
\begin{document}

\maketitle

\begin{abstract}
Data-centric distillation, including data augmentation, selection, and mixing, offers a promising path to creating smaller, more efficient student Large Language Models (LLMs) that retain strong reasoning abilities. However, there still lacks a comprehensive benchmark to systematically assess the effect of each distillation approach. This paper introduces \textbf{DC-CoT}, the first data-centric benchmark that investigates data manipulation in chain-of-thought (CoT) distillation from method, model and data perspectives. Utilizing various teacher models (e.g., \texttt{o4-mini}, \texttt{Gemini-Pro}, \texttt{Claude-3.5}) and student architectures (e.g., $3B$, $7B$ parameters), we rigorously evaluate the impact of these data manipulations on student model performance across multiple reasoning datasets, with a focus on in-distribution (IID) and out-of-distribution (OOD) generalization, and cross-domain transfer. Our findings aim to provide actionable insights and establish best practices for optimizing CoT distillation through data-centric techniques, ultimately facilitating the development of more capable reasoning models. The codebase can be accessed \href{https://github.com/UNITES-Lab/Distillation-Bench}{here}.
\end{abstract}

\section{Introduction}

\vspace{-1mm}

\begin{wrapfigure}{r}{0.50\textwidth} %
\vspace{-6mm}
    \centering
    \includegraphics[width=\linewidth]{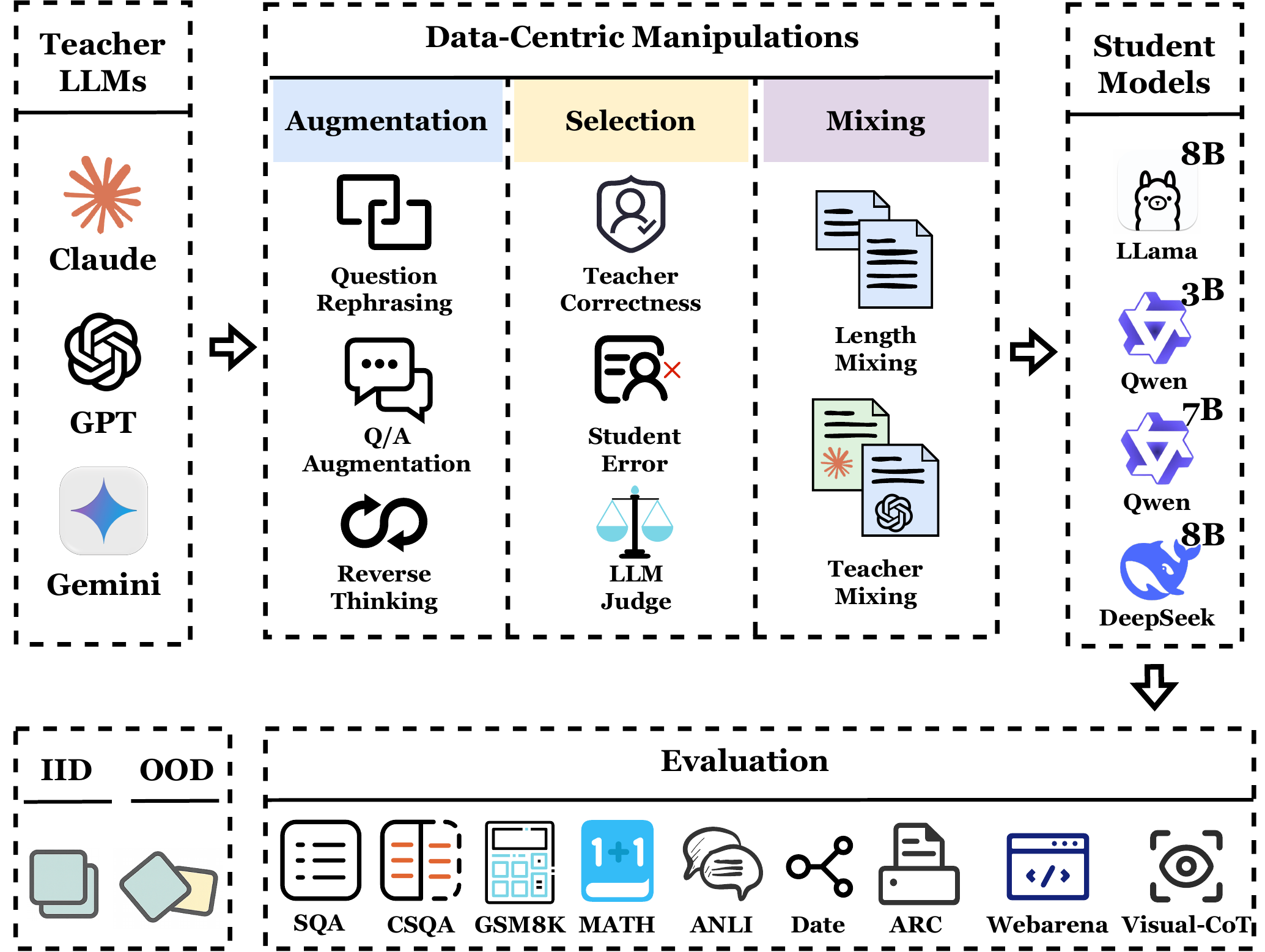} 
    \vspace{-2.5mm}
    \caption{Overview of DC-CoT pipeline. }
    \label{fig:0} 
    \vspace{-1em} 
\end{wrapfigure}

Large language models (LLMs) achieve strong reasoning performance when combined with \textit{chain-of-thought} (CoT) prompting \citep{wei2022chain}, but the best performance typically comes from expensive models with tens or hundreds of billions of parameters. To address it, \textit{knowledge distillation} (KD) stands out to transfer reasoning skills to lighter students (e.g.\ 3–8\,B) at low inference cost \citep{hinton2015distilling,ho2022large,mukherjee2023orca,wang2022self}.
Among various KD strategies for CoT~\cite{xu2024survey,tan2024large}, data-centric methods—such as augmentation, selection, and mixing—have gained popularity for being architecture-agnostic and cost-efficient~\cite{xu2023wizardlm}. However, a systematic assessment is still lacking to evaluate the effectiveness of these techniques.

To address this, building a \emph{data-centric} benchmark is essential.
Such a benchmark will provide a clearer understanding of the performance of existing data-centric methods by systematically evaluating and answering fundamental questions, such as how to effectively synthesize, select, and mix various CoT samples to robustly boost the student models' performance.
Furthermore, a data-centric benchmark will serve as a valuable and controlled evaluation resource for future research and the development of new techniques in this area.
In this work, we introduce \textbf{DC-CoT}, the first benchmark designed to investigate data-centric CoT distillation systematically, answering the following research questions:





\begin{mdframed}[userdefinedwidth=.99\linewidth,align=center,skipabove=3pt,skipbelow=3pt,innerleftmargin=6pt,innerbottommargin=6pt,innertopmargin=6pt,roundcorner=3pt,backgroundcolor=cyan!5,linecolor=gray] \noindent \ding{182} \textbf{Method Perspective}: How can various data-centric CoT distillation methods for LLMs be categorized, and what is their comparative performance in enhancing student model reasoning?

\noindent \ding{183} \textbf{Model Perspective}: How do the relative sizes and architectures of teacher and student models influence the effectiveness of data-centric CoT distillation?

\noindent \ding{184} \textbf{Data Perspective}: How do different data characteristics and settings, such as in-distribution (IID) versus out-of-distribution (OOD) data, easy-to-hard generalization, and data availability, impact the outcomes of Chain-of-Thought distillation?
\end{mdframed}

Regarding the Method Perspective, DC-CoT investigates various data manipulation strategies across three core axes:
\emph{(i) Augmentation}: Techniques like reverse reasoning and question/answer re-phrasing beyond vanilla CoT.
\emph{(ii) Selection}: Compare heuristics such as teacher-correct filtering, student-error prioritization, and LLM-based quality judges.\emph{(iii) Mixing}: Explore blending CoT data based on length, domain, and teacher origin.
To explore the Model Perspective, DC-CoT incorporates diverse teacher models (e.g., GPT-4o, Claude 3.5, Gemini-1.5-Pro) and various open-source student model families and sizes (e.g., LLaMA, Qwen, Gemma at 3-8B parameters).
To address the Data Perspective, evaluations are conducted across reasoning datasets, specifically examining performance in in-distribution (IID) and out-of-distribution (OOD) settings.

Through extensive experiments, we present key findings and insights guided by research questions across multiple perspectives.
From the Method Perspective, we find data augmentation to be generally the most effective approach and provide fine-grained analyses for each manipulation method across task types.
From the Model Perspective, we confirm the roles of compatibility and learnability, highlighting their non-trivial impact on distillation and explaining why certain teacher–student pairs may fail.
From the Data Perspective, we reveal distinct scaling behaviors across augmentation methods and quantify the generalization capabilities of student LLMs across datasets.
All these insights will help guide future research toward more effective and efficient CoT distillation paradigms.



\textbf{In Summary}, our work makes the following contributions: 
\begin{mdframed}[userdefinedwidth=.99\linewidth,align=center,skipabove=3pt,skipbelow=3pt,innerleftmargin=6pt,innerbottommargin=6pt,innertopmargin=6pt,roundcorner=3pt,backgroundcolor=yellow!5,linecolor=gray]  
\begin{enumerate} [topsep=0pt, leftmargin=15pt, itemsep=0pt]
    \item[\ding{172}] We present DC-CoT, a unified, data-centric benchmark that explores data manipulation in distillation from method, model and data perspectives. 
    \item[\ding{173}] We conduct extensive experiments across diverse teacher–student pairs, tasks, and datasets, offering the first large-scale empirical overview of CoT distillation. 
    \item[\ding{174}] We distill actionable guidelines—e.g., which augmentation boosts generalization, which filtering criterion balances quality and coverage, and when heterogeneous teacher mixtures help—thereby charting a path toward smaller yet more capable reasoning models. 
\end{enumerate}
\end{mdframed}

\vspace{-2mm}
\section{Related Works}
\vspace{-1mm}
\textbf{Reasoning in LLMs. }
Chain-of-Thought (CoT) elicits explicit intermediate reasoning steps, making LLM inference more transparent and markedly more accurate on multi-step tasks \citep{wei2022chain, kojima2022large}.  
Based on this, newer \emph{long-CoT} methods—e.g., Tree-of-Thought, iterative self-reflection, and self-correction—scale CoT by exploring multiple paths and refining answers through critique \citep{yao2023tree, madaan2023self,yu2025chain,li2025system}.

\textbf{Knowledge Distillation in LLMs.}
Knowledge distillation transfers the behaviour of a large \emph{teacher} LLM to a smaller, cheaper \emph{student}.  Beyond the original ``soft-label'' paradigm \citep{buciluǎ2006model,hinton2015distilling}, recent work treats LLM‐generated instructions, responses, and rationales as synthetic supervision for supervised fine-tuning or alignment tuning \citep{kim-etal-2023-aligning,tong2024optimizing,ouyang2022training,zhang2024balancing,wang2024bpo}.  
A particularly effective variant is \emph{reasoning} or chain-of-thought (CoT) distillation: instead of imitating only the final answer, the student is trained to follow the intermediate reasoning produced by the teacher, which has proved crucial when capacity or architectural gaps exist \citep{hsieh-etal-2023-distilling,mukherjee2023orca,lewkowycz2022solving,yu2023metamath}.  
Despite promising gains, the field still lacks principled guidance on (i) which teachers, (ii) which rationales, and (iii) what selection or mixing strategies yield maximal benefit for a given student, motivating a more data-centric exploration of CoT distillation. More detailed related work is given in Appendix \ref{app:related}.

\vspace{-2mm}
\section{Methodology: A Data-Centric CoT Distillation Benchmark}
\vspace{-1mm}

\subsection{Data-centric Manipulation} \label{sec:method}

\begin{wrapfigure}{r}{0.5\textwidth} %
\vspace{-5mm}
    \centering
    \includegraphics[width=\linewidth]{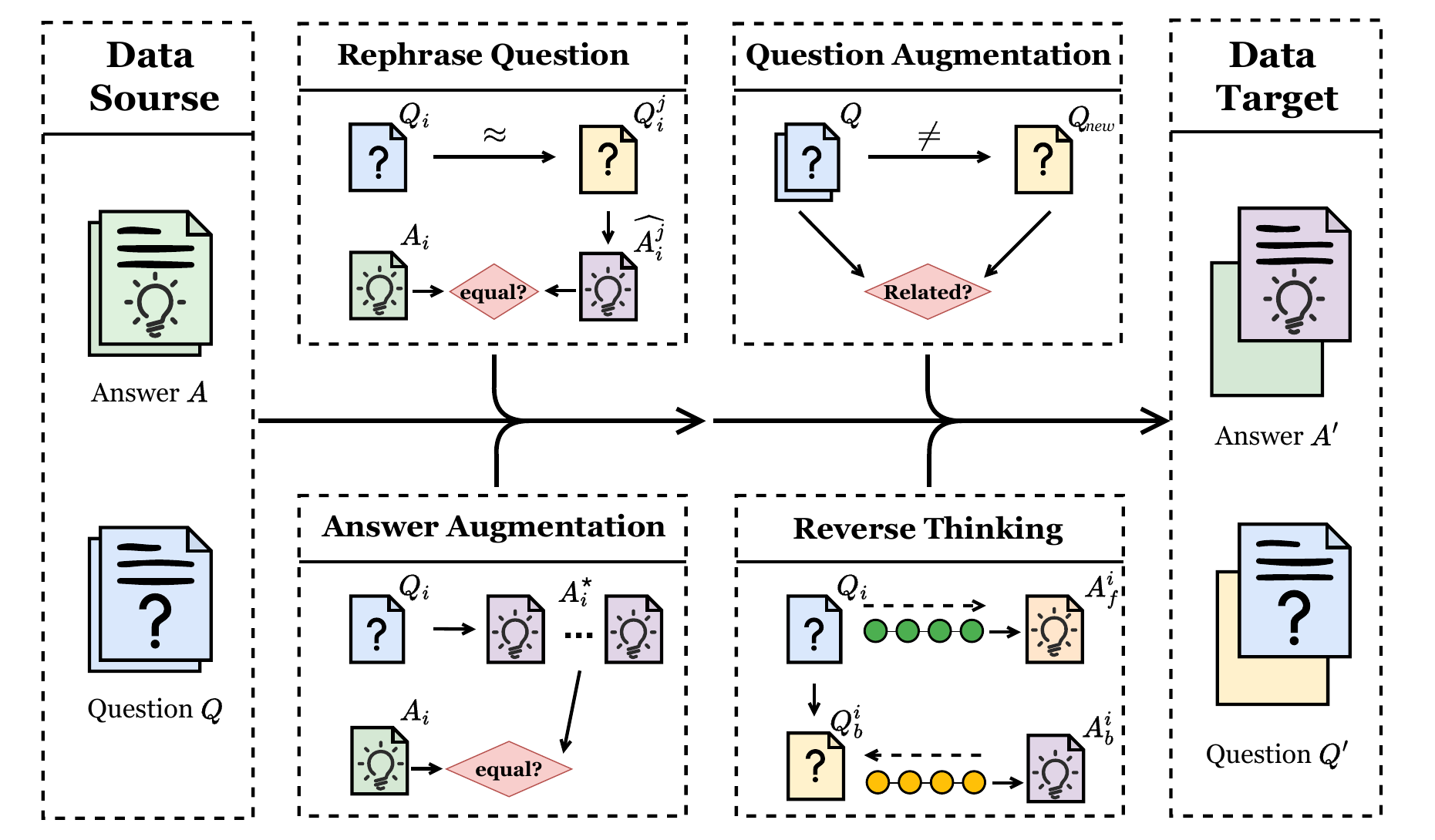} 
    \vspace{-3.5mm}
    \caption{Data-centric augmentation flow.
Teacher CoT traces are independently transformed by four operations: Rephrase Question, Question Augmentation, Answer Augmentation, and Reverse Thinking.}
    \label{fig:0} 
\end{wrapfigure}


The central theme of our \textbf{DC-CoT} is the systematic evaluation of \textit{data-centric manipulations} applied to CoT exemplars for knowledge distillation. These manipulations encompass various strategic operations to transfer the initial dataset $D^{source}$ to the target dataset $D^{target}$ for small student model training, potentially guided by a set of parameters or rules $\Theta$: $D^{target} = \mathcal{M}(D^{source}, \Theta)$. Here, $\mathcal{M}$ represents the abstract data transformation function encompassing augmentation, selection, and mixing. For augmentation strategies, we denote $L$ as the number of synthetic samples generated per source instance. \textbf{DC-CoT} is designed to deconstruct and analyze the impact of instantiating $\mathcal{M}$ through three primary types of data-centric operations: \ding{182} Data Augmentation (Section ~\ref{sec:data_augmentation}), \ding{183} Data Filtering (Section ~\ref{sec:data_filtering}), and \ding{184} Data Mixing (Section ~\ref{sec:data_mixing}).

\subsubsection{\textcolor{orange!80}{Data Augmentation}} \label{sec:data_augmentation}


\textcolor{orange!100}{Data Augmentation} is crucial in CoT distillation by enriching and diversifying the training data $(D^{source})$ available for the student model, to expose the student to various reasoning patterns, question formulations, and explanatory styles for enhancing their reasoning capabilities and generalization. Within the DC-CoT benchmark, we investigate several data augmentation strategies as follows:


\noindent\textbf{\ding{182} Question Rephrasing:} This method, introduced in MetaMath \citep{yu2023metamath}, aims to increase question diversity by having the teacher LLM $\mathcal{T}$ paraphrase an existing question $Q_i$ while preserving its underlying meaning and original answer $A^*_i$: $\{\hat{Q}_i^j = \mathcal{T}(Q_i, P_{reph})\}_{j=1}^{L}$. Here $Q_i$ and $P_{reph}$ are the original question and rephrasing prompt. For each rephrased question \( \hat{Q}_i^j \), the teacher \( \mathcal{T} \) generates a CoT rationale \( \hat{R}_i^j \) and answer \( \hat{A}_i^j \). one augmentation is retained if \( \hat{A}_i^j \) matches the original answer.


\noindent\textbf{\ding{183} Question Augmentation:} This strategy focuses on creating entirely new questions $Q_{new}$, to broaden the topical coverage or complexity of the training data, based on a set of seed questions $Q$~\citep{li2024common}: $Q_{new} = \mathcal{T}(Q, P_{QA})$. $P_{QA}$ here is a prompt for generating novel questions. After that, the same generation-then-filter process will be adopted to produce new answers and CoTs for the augmented questions, as we introduced in the Question Rephrasing method. Unlike general instruction-tuning methods (e.g., Self-Instruct), this operation is strictly constrained to \textit{Reasoning Transfer}. The prompt $P_{QA}$ forces the generation of parallel reasoning problems (e.g., altering numerical values in math or subjects in logic puzzles) to ensure the student learns the underlying reasoning pattern rather than memorizing specific answers.

\noindent\textbf{\ding{184} Answer Augmentation:} It involves prompting the teacher LLM $\mathcal{T}$ to generate multiple diverse CoT rationales $R$ that all lead to the same correct ground-truth answer $A_i^*$~\citep{yu2023metamath}. Given $(Q_i, A_i^*)\in D^{source}$, and using a CoT generation prompt $P_{AA}$, the teacher model generates $L$ candidate rationales and answers as follows: $\{(R_i^k, A_i^k) = \mathcal{T}(Q_i, P_{AA}, \text{temp})\}_{k=1}^L$. To mitigate the risk of reasoning hallucinations, the prompt explicitly conditions the teacher on the ground-truth answer $A^*_i$. Our empirical results suggest that the benefit of exposing the student to diverse valid reasoning paths outweighs the noise of occasional imperfect traces, as the student learns the intersection of valid logic across the augmented set.

\noindent\textbf{\ding{185} Reverse Thinking Augmentation}
Reverse Thinking was introduced in the RevThink \citep{chen2024reverse}. The goal is to enrich the data by generating forward CoT reasoning $R_f$, a corresponding backward question $Q^b$, and backward reasoning $R_b$. For each $(Q_i, A_i)\in D^{source}$ we do the following:

\begin{itemize}[leftmargin=*,itemsep=1pt,topsep=2pt,parsep=1pt]
    \vspace{-2mm}
    \item \textit{Generate Forward Reasoning:} $R_f^i=\mathcal{T}(Q_i, P_f)$ for some prompt $P_f$. This is filtered to ensure that the outcome of $R_f^i$ is the ground truth $A_i$.
    \item \textit{Generate Backward Question:} Using a prompt $P_{bq}$, the teacher $\mathcal{T}$ generates a question that inverts the original problem: $Q_b^i=\mathcal{T}(Q_i, A_i, P_{bq})$. 
    \item \textit{Generate Backward Reasoning:} The teacher then generates the CoT for this backward question: $R_b^i=\mathcal{T}(Q_b^i, P_{br})$ for some prompt $P_{br}$.
    \item \textit{Consistency Filtering:} A consistency check $c=\mathcal{T}(Q_i, A_i, Q_b^i, R_b^i, P_{con})$ is performed for making sure the backward and the forward questions are related and consistent with each other~\cite{yang2025quantifying}. Only consistency quadruplets $(Q_i, R_f^i, Q_b^i, R_b^i)$ where $c=1$ are retained.
\end{itemize}

\vspace{-4mm}
\subsubsection{\textcolor{purple!80}{Data Filtering}} \label{sec:data_filtering}
\vspace{-1mm}

\begin{wrapfigure}[17]{r}{0.5\textwidth} %
\vspace{-3mm}
    \centering
    \includegraphics[width=\linewidth]{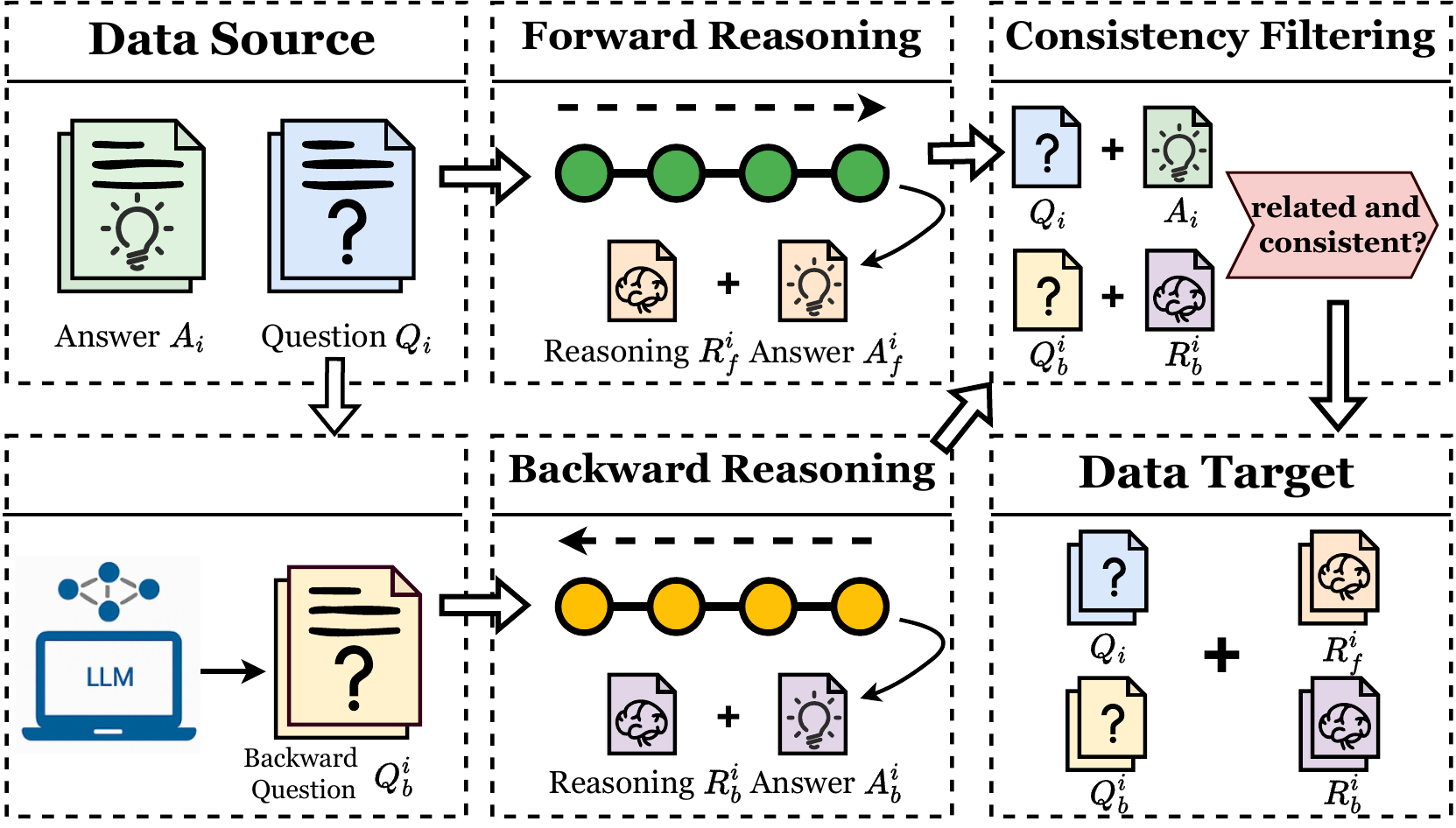} 
    \vspace{-4mm}
    \caption{Reverse-Thinking augmentation pipeline: from each (question, answer) pair, generate forward reasoning, synthesize a backward question with its reasoning, then keep only examples whose forward-backward chains pass a consistency check.}
    \label{fig:4} 
    \vspace{-3mm} 
\end{wrapfigure}

\textcolor{purple!100}{Data Filtering}, or selection, is a critical step applied to either initial source data $D^{source}$ or augmented data to create a high-quality training set $D^{train}$ for the student model. Since not all CoT instances are equally beneficial, as some are noisy or incorrect, filtering aims to identify and retain the most valuable exemplars to optimize learning. Our DC-CoT investigates the following data selection strategies: 


\noindent\textbf{\ding{182} Filtering by Teacher Correctness:} This strategy used in \citep{ho2022large}, retains CoT instances where the teacher's final answer $A_i$ matches the ground-truth answer $A_i^*$: $D^{target} = \{(Q_i, R_i, A_i) | A_i = A^*_i\}$. This ensures the student learns from CoTs lead to correct outcomes. 


\noindent\textbf{\ding{183} Filtering by Student Error:} 
This filtering strategy focuses student learning on its weaknesses by selecting instances where the student model yields an incorrect answer: $D^{target} = \{(Q_i, R_i, A_i) | \hat{A}_i \neq A_i^*\}$. This concentrated learning can focus on students' underperformed areas. 


\noindent\textbf{\ding{184} LLM-as-a-Judge Filtering:}
Inspired by I-SHEEP \citep{liang2024sheep}, this method uses an external LLM $\mathcal{L}_{\text{judge}}$ to assess CoT instance quality based on criteria like coherence, correctness, and clarity, allowing for a nuanced quality assessment~\cite{li2024generation,li2025preference}: $\text{Score}_i= \mathcal{L}_{\text{judge}}(A_i, R_i, Q_i, P_{eval})$. Instances are retained if their score meets a threshold $\tau$, making the final dataset become: $D^{source} = \{(Q_i, R_i, A_i) | \text{Score}_i \geq \tau\}$. To validate the reliability of this automated judge, we conducted a human evaluation on a random sample of 100 filtered instances from SQA and GSM8K. We observed a Cohen’s Kappa ($\kappa$) of 0.84, indicating strong agreement between the LLM Judge (GPT-4o) and human experts, with the Judge exhibiting a slight preference for strictness—a desirable bias for high-quality distillation.

\begin{wrapfigure}[15]{r}{0.5\textwidth}
\vspace{-4mm}
    \centering
    \includegraphics[width=\linewidth]{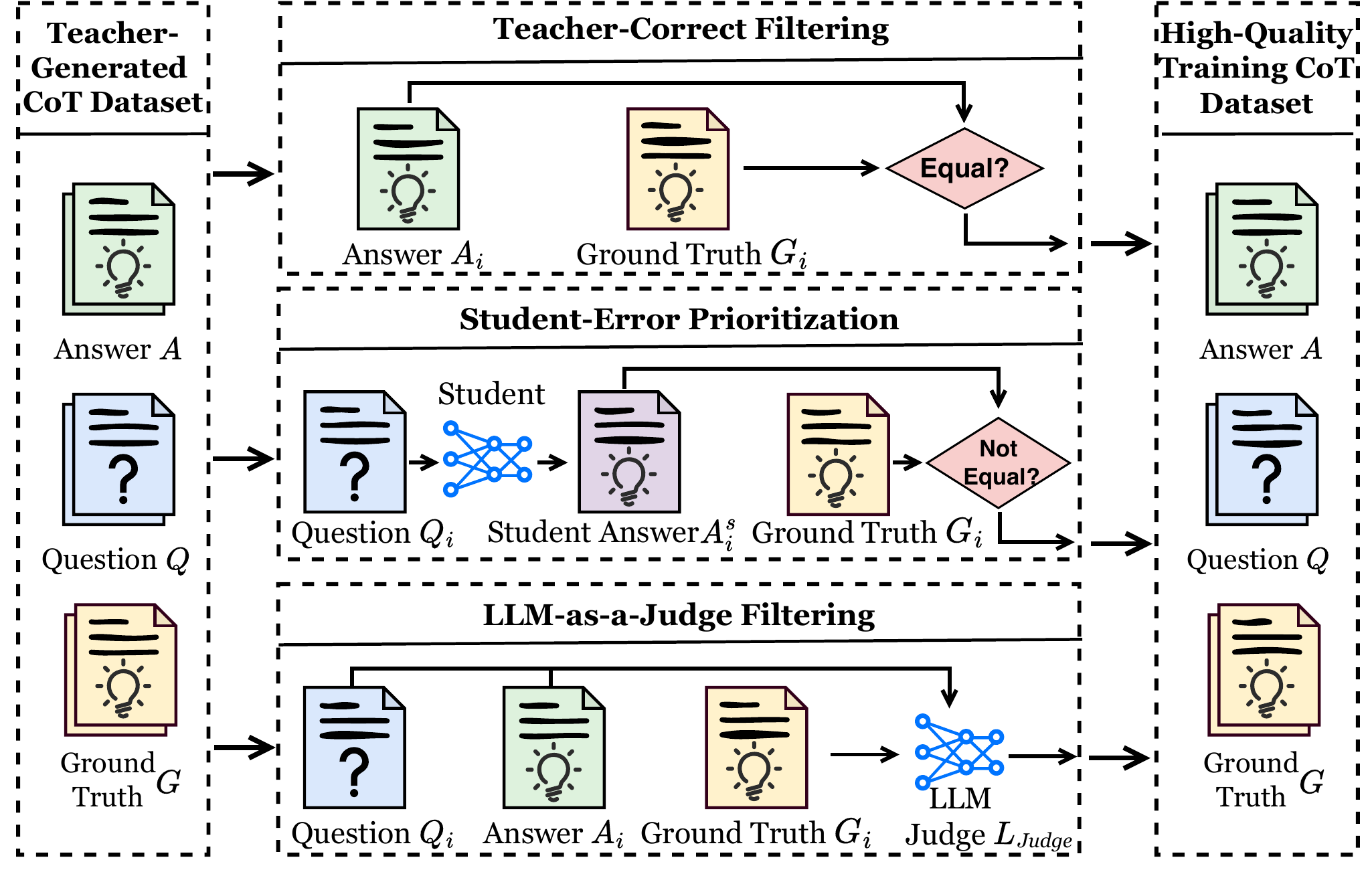} 
    \caption{Data-filtering pipeline in DC-CoT. A teacher-generated CoT pool is refined through three selectors.
    }
    \label{fig:3} 
\end{wrapfigure}
\subsubsection{\textcolor{cyan!80}{Data Mixing}} \label{sec:data_mixing}
\vspace{-1mm}
Beyond augmentation and selection, \textcolor{cyan!100}{Data Mixing} offers another avenue for data-centric manipulation in CoT distillation. This strategy involves strategically combining CoT instances from different distributions or with varying characteristics to create a more diverse training dataset $D^{target}$ for the student model. The core idea is that a blend of reasoning styles, complexities, or teacher provenances can lead to a student model with more robust and generalizable reasoning capabilities.


\noindent\textbf{\ding{182} Length-based CoT Mixing:} Length-based mixing, introduced in \citep{li2025small}, combines CoT examples of varying reasoning length to help bridge this learnability gap for smaller models and offers complexity for larger models. This mix, controlled by a ratio $\alpha$, aims to provide a balanced curriculum, exposing students to detailed and concise reasoning.


\noindent\textbf{\ding{183} Teacher-based CoT Mixing:} This method blends CoTs generated by different teachers~\citep{li2025small}. The mixed dataset is again guided by a ratio $\alpha$, providing a balanced set of reasoning examples and preventing smaller students from being overwhelmed while still offering sophisticated examples.

\vspace{-2mm}
\section{Experiment Result \& Analysis}
\vspace{-1mm}
\subsection{Benchmark Setup}
\noindent \textbf{Teacher Models.} We use SoTA LLMs known for strong reasoning to generate CoT rationales: (1) Gemini-1.5-Pro \citep{team2024gemini}, (2) GPT-4 \citep{achiam2023gpt}, (3) Claude-3.5 Sonnet \citep{claude3.5sonnet2024}, (4) GPT-4.1 mini \citep{gpt4.1openai2024}, (5) o4 mini \citep{o3o4mini2024}. Utilizing multiple teachers allows us to study the impact of teacher diversity. Data filtering is performed with task-specific Judge LLMs: LLama-2-70B for textual tasks, GPT-4o-mini \citep{achiam2023gpt} for agentic tasks, and GPT-4/4.1-mini for visual tasks.
\vspace{-2mm}

\noindent \textbf{Student Models.} We test these open-source models as students: (1) LLama-3.1-8B \citep{grattafiori2024llama}, (2) LLama-3.1-8B-R1 Distilled \citep{guo2025deepseek}, (3) Mistral-7B \citep{Jiang2023Mistral7}, (4) Gemma-7B \citep{team2024gemma}, and (5) Qwen-2.5-7B \citep{yang2024qwen2}. 
\noindent \textbf{Baselines.} For Baseline comparison, we evaluate the models for (1) Zero Shot performance on the tasks, (2) Generate Zero-Shot CoT \citep{kojima2022large}, (3) Fine-tune the model on the dataset without any CoT, and (4) Vanilla CoT generated by the teacher model with no augmentation/filtering/mixing.

\vspace{-2mm}
\noindent \textbf{Datasets.} Student performance is assessed on diverse reasoning datasets covering various skills and complexities. We evaluate textual reasoning tasks on: \textit{Commonsense Reasoning Tasks:} StrategyQA (SQA; \citep{geva-etal-2021-aristotle}), CommonsenseQA (CSQA; \citep{talmor-etal-2019-commonsenseqa}), ARC-challenge (ARC; \citep{clark2018think}). \textit{Math Reasoning:} GSM8K (GSM8K; \citep{cobbe2021training}), MATH (MATH; \citep{hendrycks2021measuring}). \textit{Natural Language Inference:} ANLI (ANLI; \citep{nie-etal-2020-adversarial}). \textit{Logical Reasoning:} Date Understanding (Date; \citep{srivastava2022beyond}). We evaluate agentic reasoning tasks on \textsc{WebArena} \citep{zhou2023webarena}, and evaluate visual reasoning on Visual-CoT \citep{shao2024visual}, OK-VQA \citep{marino2019ok}, and CLEVR \citep{johnson2017clevr}. We classify Shopping, Map, and Reddit as webarena-easy, and others as hard. For task descriptions, please refer to ~\ref{app:tasks}.





\begin{table*}[]
\small
\caption{Accuracy of augmentation, selection and mixing strategies on Llama-3.1-8B. Textual scores are the mean of three independent runs.}
\centering
\resizebox{\textwidth}{!}{  
\begin{tabular}{lccccccccccccc}
\toprule[1.2pt]
& Agentic &  \multicolumn{3}{c}{Visual} & \multicolumn{7}{c}{Textual}           & \multicolumn{2}{c}{AVG.} \\  \cmidrule(lr){2-2} \cmidrule(lr){3-5} \cmidrule(lr){6-12} \cmidrule(lr){13-14}
& WebArena             & Visual-CoT &OK-VQA	&CLEVR           & SQA & CSQA & ARC & MATH & GSM8K & ANLI & Date   & Visual+Agentic  &    Textual                 \\ \hline 
\multicolumn{14}{c}{{\emph{\textbf{Data Augmentation}}}}\\\midrule       
\rowcolor{orange!15}
Zero Shot             &          5.66           &      42.10   &\textbf{65.60}	&56.88       &  57.64   &  43.08    &  48.46   &  9.32    &  19.64     & 33.83       &  49.70 & 23.88    &        37.38        \\
\rowcolor{orange!15}
Zero Shot CoT             &       8.25           &        44.52  & 61.84	& \textbf{58.36 }     &   65.55  &  53.56    &  67.41   & 11.76     & 21.00      &  39.92      &  62.13   & 26.39  &         45.90       \\
\rowcolor{orange!15}
No CoT                &           \textbf{30.05}           &       \textbf{46.66} & 62.18 & 52.12     &  59.89   &  65.36    &  60.41   &  7.39    &  20.74     &  35.42      & 50.37  & \textbf{38.36}  &            42.80       \\
\rowcolor{orange!15}
Vanilla CoT           &           22.78           &    45.44  & 59.94	& 54.04     &   58.08  &  69.37    &  55.63   &  4.38    &  24.30     & 23.92      &  57.02  & 34.11 &           41.81       \\
\rowcolor{orange!15}
Rephrase Question &            -          &        -  &            -          &        -           & 59.73    &  62.95    &  67.01   &   16.52   &  38.86     &  42.47      &    59.41    & -   &       49.56     \\
\rowcolor{orange!15}
Question Aug &            -      &            -          &        -         &        -        &  60.40   &  61.47    &   70.37  &  20.31    &  44.03     &    41.26    &   61.07      & -   &       51.27     \\
\rowcolor{orange!15}
Answer Aug &           -      &            -          &        -          &         -       &  64.49   &  64.57    &  81.61   &  \textbf{36.84}    &   53.48    &    40.29    &   61.80    & -   &     57.58         \\
\rowcolor{orange!15}
Reverse Thinking     &     &            -          &        -          -            &        -        &   \textbf{72.49}  &   \textbf{78.46}   &  \textbf{82.17}   &  35.52    &  \textbf{76.35}     & \textbf{49.75}       &  \textbf{70.41}   & -  &          \textbf{66.45}       \\ \hline 
\multicolumn{14}{c}{{\emph{\textbf{Data Selection}}}}\\\midrule   
\rowcolor{purple!15}
No Selection          &            22.78          & 44.52  & 59.94	& 54.04 &  59.89   &  65.36    &  60.41   &  \textbf{7.39}    &  20.74     &  \textbf{35.42}      & 50.37  &33.65 &         42.80
           \\
           \rowcolor{purple!15}
Filtering with Teacher     &           14.66           &       45.50    &63.80	&\textbf{67.60}     &  \textbf{61.43}     &   70.72   &  \textbf{62.86}   &  5.04    &  \textbf{30.27}    &  24.11      &  58.69     & 30.08  &       \textbf{44.73} 

       \\
       \rowcolor{purple!15}
Filtering with Student     &          \textbf{27.59}            &        45.90    & \textbf{66.30}	& 57.02    &  60.29   &   \textbf{70.85}   &  60.30   &  5.21    & 26.97      &   25.40     &   58.04    & \textbf{36.75}  &       43.87  

       \\
       \rowcolor{purple!15}
Judge LLM        &        15.64              &   \textbf{46.54} & 59.42	& 54.12 &   54.83        &  62.49   &  57.46    & 3.43    &   22.72   &  26.51     &  \textbf{59.85}   & 31.09    &        41.04 

               \\ \hline 
 \multicolumn{14}{c}{{\emph{\textbf{Data Mixing}}}}\\\midrule 
 \rowcolor{cyan!15}
No Mixing             &              \textbf{22.78}          & 44.52  & 59.94	& 54.04  &  \textbf{59.89}   &  65.36    &  60.41   &  \textbf{7.39}    &  20.74     &  \textbf{35.42}      & 50.37 & \textbf{33.65} & \textbf{42.80} 
\\
\rowcolor{cyan!15}
Length Mixing         &          -      &          -            &   -     &   -  &   58.58        & \textbf{68.04}    & 54.79     &  4.64   &  \textbf{21.84}    &   22.50    &  \textbf{59.63}    & -   &          41.43            \\
\rowcolor{cyan!15}
Teacher Mixing        &          21.18            &    \textbf{45.48} & \textbf{61.7} &	\textbf{55.6}  & 56.75        &  66.94   & \textbf{62.82}    &  5.96   &   19.57   & 29.46      &    52.30    & 33.33 &         41.97             \\ \toprule[1.2pt]                 
\end{tabular}}
\label{tab:1}
\vspace{-7mm}
\end{table*}
\subsection{Method-Level Results}

This section delves into the performance of various data-centric manipulation strategies by posing key questions and deriving insights from our experimental findings. The analysis primarily references Table~\ref{tab:1}. It is important to note that the results discussed in Table~\ref{tab:1} all pertain to the \textit{Llama-3.1-8B} student model. Furthermore, the teacher model for visual tasks was \textit{GPT-4-mini} \citep{achiam2023gpt}, for agentic tasks it was \textit{Claude-3.5} \citep{claude3.5sonnet2024}, and for textual tasks, \textit{Gemini-1.5-Pro-001} \citep{team2024gemini} was used. For the mixing, we use the models as described in Table~\ref {tab:2} and~\ref {tab:3}.

\noindent \textbf{Q1: How do the broad categories of data-centric manipulation compare in terms of overall effectiveness?} Table~\ref{tab:1} shows that Data Augmentation strategies yield the most substantial average performance uplift over the Vanilla CoT baseline. For instance, Reverse improves average accuracy on all eight tasks by $24.64\% \textcolor{pygreen}{\uparrow}$. Filtering with Teacher Correctness (Textual Average: $44.7\%$) improves by $+1.93 \textcolor{pygreen}{\uparrow}$ over Vanilla CoT. The best mixing strategy, Teacher Mixing (Textual Average: $41.97\%$), shows a marginal decrease of $0.83\%\textcolor{red}{\downarrow}$ over Vanilla CoT. This confirms that for a moderately sized student (7-8B), creating diverse rationales is more impactful than selecting or reshuffling existing ones. Data selection is vital for quality control, and data mixing helps tailor its composition.

\begin{wrapfigure}{r}{0.48\textwidth}
    \centering
    \small
    \vspace{-4mm}
    \captionof{table}{Comparison of Data-Centric Distillation vs. Logit-based KD on ARC-Challenge (Teacher: Llama-3.1-70B).}
    \vspace{-2mm}
    \resizebox{\linewidth}{!}{%
    \begin{tabular}{llc}
        \toprule[1.2pt]
        Method & Access Required & Accuracy (\%) \\
        \hline \addlinespace[2pt]
        Teacher Baseline & Weights/Logits & 92.4 \\
        \hline \addlinespace[2pt]
        Standard KD (KL Div.) & Weights/Logits & 64.8 \\
        Vanilla CoT (SFT) & Black-box (Text) & 60.4 \\
        \textbf{DC-CoT (Reverse)} & \textbf{Black-box (Text)} & \textbf{69.2} \\
        \toprule[1.2pt]
    \end{tabular}}
    \vspace{-4mm}
    \label{tab:logit_vs_data}
\end{wrapfigure}
\textbf{Comparison with Logit-based Distillation.} While DC-CoT focuses on black-box distillation (where teacher logits are unavailable), we assessed its competitiveness against white-box methods using an open-weights teacher (Llama-3.1-70B) on the ARC-Challenge. As shown in Table~\ref{tab:logit_vs_data}, DC-CoT (Reverse Thinking) achieved $69.2\%$, significantly outperforming standard Logit-based KD $64.8\%$. This suggests that transferring explicit reasoning steps via data augmentation is more effective for reasoning tasks than minimizing divergence on the output distribution alone.

\noindent \textbf{Q2: Which techniques are most effective for each data manipulation?} From Table~\ref{tab:1}, Reverse consistently excels, especially for structure logical deduction (MATH, GSM8K, Date). It likely fosters a deeper understanding by teaching bi-directional reasoning. Answer Augmentation also performs robustly, particularly for commonsense reasoning (SQA, CSQA), by exposing the student to varied solution paths, enhancing flexibility. While Question Augmentation and Rephrasing increase diversity, the more profound alterations from Reverse and Answer Augmentation generally yield larger gains. Among the selection techniques, LLM-as-a-Judge filtering is highly effective, often surpassing simpler heuristics due to its nuanced assessment of rationale quality (coherence, soundness) beyond mere answer correctness. However, filtering by Teacher Correctness is a strong baseline, ensuring students learn from factually accurate paths and consistently improve over no selection or other methods. When compared to the \textit{No Mixing} baseline, data mixing strategies show varied effects. Length Mixing (Average: $41.43\%$) results in a slight decrease of $1.37\%\textcolor{red}{\downarrow}$ on average for textual tasks. However, while underperforming on others, it shows improvements on specific textual datasets like CSQA, GSM8K, and Date. Teacher Mixing also shows a slight decrease of $0.83\%\textcolor{red}{\downarrow}$ on average for textual tasks compared to \textit{No Mixing}. These results suggest that the benefits of the tested mixing strategies are not universally additive over a strong \textit{No Mixing} baseline for textual tasks on average, but they can offer advantages for specific datasets or modalities, likely by tailoring the data complexity or teacher style to particular student needs or task characteristics.

\begin{table*}[]
\small
\centering
\caption{Reverse-augmented distillation results for different teacher / student combinations on textual tasks; numbers are three-run averages.}
\resizebox{0.85\textwidth}{!}{
\begin{tabular}{llcccccccr}
\toprule[1.2pt]
Student Model                             & Teacher Model       & SQA & CSQA & ARC & MATH & GSM8K & ANLI & Date & AVG. \\ \hline \addlinespace[2pt]
\multirow{2}{*}{Llama-3.1-8B}               & Gemini-1.5-Pro &      \textbf{72.49}  & \textbf{78.46}  & 82.17  & 35.52  & \textbf{76.35}  &  49.75  & \textbf{70.41}  & \textbf{66.45}\\
& GPT-4              &  70.74    &   71.93   &  \textbf{83.64}   &  34.60    &   70.72    &   \textbf{51.37}     &   68.51  & 64.50
  \\ \hline \addlinespace[2pt]
\multirow{2}{*}{Llama-3.1-8B-R1}    & Gemini-1.5-Pro      & 69.43  & 71.74 & 74.23 & \textbf{36.82}  & 69.45  & 47.08  & \textbf{70.41}  &  62.74
\\
& GPT-4               &  70.95   &  68.40    &  76.84   &  36.27    &   70.94    &    50.58    &  67.80  & 63.11
 \\ \hline \addlinespace[2pt]
\multirow{2}{*}{Mistral-7B} & Gemini-1.5-Pro      &  72.05  & 75.53  & 76.96  & 16.12  & 59.21  & 45.00  & 59.17  & 57.72 \\
& GPT-4               &  71.08   &  72.63    &  76.85   & 15.39     &  58.86     &   45.62     &  60.19  &  57.23
 \\ \hline \addlinespace[2pt]
\multirow{2}{*}{Gemma-7B}  & Gemini-1.5-Pro      &  68.12  & 74.86  & 73.46  & 16.54  & 53.45  & 40.92  &  31.36      &51.24 \\
& GPT-4               &  69.08   & 73.81     &  75.60   & 16.18     &   54.49    &    41.65    &   30.57   &   51.63 \\\toprule[1.2pt]
\end{tabular}}
\label{tab:2}
\vspace{-8mm}
\end{table*}

\noindent \textbf{Q3: Which data-centric methods show particular strengths for specific reasoning tasks?}Optimal strategies vary by task demands, and combining effective augmentation with suitable filtering or mixing can yield further improvements:
\begin{enumerate}[leftmargin=*,itemsep=1pt,topsep=2pt,parsep=1pt]
    \item \textit{Textual Reasoning (SQA, CSQA, ANLI):} Answer Augmentation and Question Rephrasing enhance linguistic diversity. These should be combined with LLM-as-a-Judge filtering to ensure the high quality and coherence of the textual rationales. Teacher Mixing could also be beneficial after augmentation for tasks with varying teacher capabilities.
    \item \textit{Mathematical Reasoning (GSM8K, MATH, Date):} Reverse Thinking excels due to the need for backward deduction. Answer Augmentation is also valuable. These augmented datasets should then be rigorously filtered using Filtering by Teacher Correctness to eliminate any incorrect mathematical procedures. Subsequently, Length Mixing can be applied to balance the complexity of CoTs presented to the student. 
    \item \textit{Agentic Reasoning (WebArena):} Given the complexity and potential for action chain errors, the augmented data should be curated using LLM-as-a-Judge filtering to enhance correctness. 
    \item \textit{Visual Reasoning (Visual-Cot):} It is critical to use LLM-as-Judge filtering to ensure rationales are not only logically sound but also accurately reflect and reference the visual content. 
\end{enumerate}

\subsection{Model-Level Results}
We explored the effect of Teacher and Student types/sizes as well. For detailed results on Student Models, please refer to Appendix~\ref{app:student}.
\subsubsection{Teacher Model Analysis}
We investigate the interplay between teacher and student models, summarized in Tables~\ref{tab:2} and~\ref{tab:3}. For textual reasoning tasks, we utilize the best-performing augmentation approach, Reverse, and for visual as well as agentic tasks, we report the performance on vanilla CoT. 

\begin{wrapfigure}[10]{r}{0.45\textwidth}
\vspace{-5mm}
    \centering
    \small
    \captionof{table}{Impact of teacher model on agentic (WebArena) and visual (Visual-CoT) performance.}
    \scalebox{0.8}{
    \begin{tabular}{llcc}
        \toprule[1.2pt]
        Student Model             & Teacher Model & WebArena          & Visual-CoT                 \\ \hline \addlinespace[2pt]
        \multirow{2}{*}{Llama-3.1-8B} & Claude-3.5    & 22.78             & -                          \\
                                 & GPT-4o        & \textbf{24.51}    & -                          \\
        \hline \addlinespace[2pt]
        \multirow{2}{*}{Llama-3.1-8B-R1} & Claude-3.5    & 11.33             & -                          \\
                                   & GPT-4o        & 13.79             & -                          \\
        \hline \addlinespace[2pt]
        \multirow{3}{*}{Qwen-2.5-VL-3B}
                                   & GPT-4         & -                 & 42.92                      \\
                                   & GPT-4-mini    & -                 & \textbf{45.44}             \\
                                   & o4-mini       & -                 & 45.20                      \\ \toprule[1.2pt]
    \end{tabular}}
    \vspace{-2mm}
    \label{tab:3}
\end{wrapfigure}
\noindent\textbf{Q4. How does the choice of a teacher model impact the performance of different student models on textual reasoning tasks? Is there a universally ``best'' teacher for all students?} Table~\ref{tab:2} reveals that for textual reasoning, stronger models like Gemini-1.5-Pro and GPT-4 generally yield better results when distilling to capable student models such as Llama-3-8.1 B. For instance, Llama-3.1-8B achieves a high average textual score for both teachers, suggesting that as long as the teacher is powerful enough and the student has adequate capacity, transferring complex reasoning using Knowledge Distillation is quite effective. However, a universally ``best'' teacher is not apparent. While Gemini-1.5 shows a slight edge for LLama-3.1-8B on average, GPT-4 can be comparable or better on specific datasets (e.g., ARC for Llama-3.1-8B). For Mistral Gemini-1.5, it slightly outperforms GPT-4, whereas for Gemma-7B, GPT-4 is marginally better than the other. This variability indicates that optimal teacher-student pairings are nuanced, likely influenced by factors like architectural alignment or specific knowledge domains. 

\begin{figure*}[t!] 
\vspace{-3mm}
    \centering
    \includegraphics[width=0.9\textwidth]{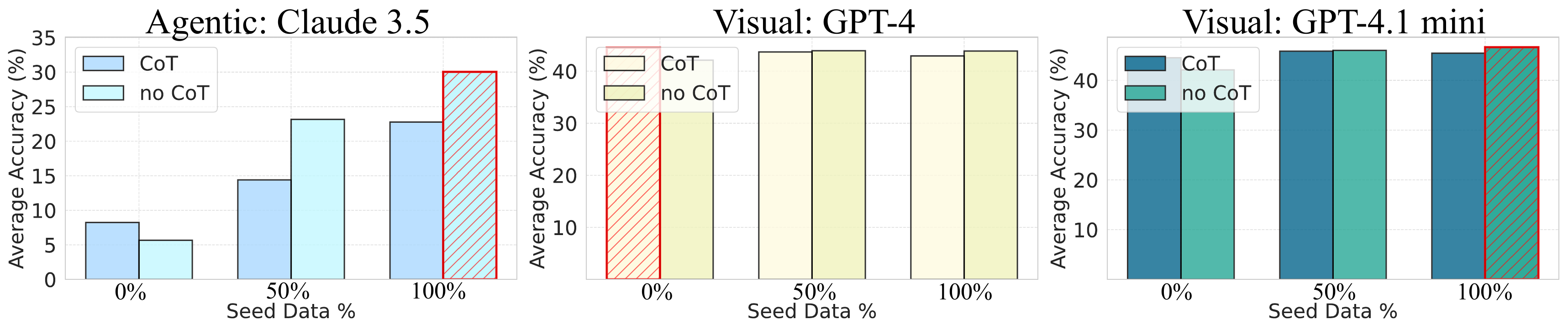}
    \vspace{-2mm}
    \caption{Accuracy of different seed-data sizes and teachers for WebArena and Visual-CoT.}
    \label{fig:bar} 
    \vspace{-9mm}
\end{figure*}

\begin{wrapfigure}{r}{0.57\textwidth}
\vspace{-4mm}
    \centering
    \small
    \captionof{table}{Performance of Llama-3.1-8B and Mistral-7B when varying the percentage of seed data.}
    \resizebox{\linewidth}{!}{
    \begin{tabular}{lllccccr}
        \toprule[1.2pt]
        Student Model             & Seed Data \% & Augmentation Type & SQA   & ARC   & GSM8K & Date  & AVG.   \\ \hline \addlinespace[2pt]
        \multirow{10}{*}{Llama-3.1-8B} & \multirow{1}{*}{Zero-Shot} & None            & 57.64 & 48.46 & 19.64 & 49.70 & 43.86  \\
                                     \arrayrulecolor{gray}\cmidrule(lr){2-8}
                                     & \multirow{2}{*}{25\%}      & Vanilla CoT     & 68.12 & 79.95 & 42.99 & 66.86 & 64.48  \\
                                     &                             & Reverse         & 60.70 & 77.82 & 30.02 & 74.56 & 60.78  \\
                                     \arrayrulecolor{gray}\cmidrule(lr){2-8}
                                     & \multirow{2}{*}{50\%}      & Vanilla CoT     & \textbf{73.80} & 80.12 & 36.39 & 71.01 & 65.33  \\
                                     &                             & Reverse         & 62.88 & 79.95 & 47.01 & 68.64 & 64.62  \\
                                     \arrayrulecolor{gray}\cmidrule(lr){2-8}
                                     & \multirow{2}{*}{75\%}      & Vanilla CoT     & 67.69 & 71.78 & 26.61 & 65.89 & 57.99  \\
                                     &                             & Reverse         & 68.12 & 80.79 & 59.67 & \textbf{73.96} & 70.64  \\
                                     \arrayrulecolor{gray}\cmidrule(lr){2-8}
                                     & \multirow{2}{*}{100\%}     & Vanilla CoT     & 58.08 & 55.34 & 24.30 & 59.41 & 49.28  \\
                                     &                             & Reverse         & \textbf{72.49} & \textbf{82.17} & \textbf{76.35} & 70.41 & \textbf{75.36} \\
        \toprule[1.5pt]
        \multirow{10}{*}{Mistral-7B} & \multirow{1}{*}{Zero-Shot} & None            & 55.02 & 50.94 & 20.24 & 46.75 & 43.24  \\
                                     \arrayrulecolor{gray}\cmidrule(lr){2-8}
                                     & \multirow{2}{*}{25\%}      & Vanilla CoT     & 70.46 & 69.52 & 44.09 & 63.58 & 61.91  \\
                                     &                             & Reverse         & 71.18 & 73.98 & 54.13 & 62.72 & 65.50  \\
                                     \arrayrulecolor{gray}\cmidrule(lr){2-8}
                                     & \multirow{2}{*}{50\%}      & Vanilla CoT     & 64.91 & 70.04 & 39.25 & 58.53 & 58.18  \\
                                     &                             & Reverse         & 68.56 & 76.11 & 53.90 & \textbf{64.59} & 65.79  \\
                                     \arrayrulecolor{gray}\cmidrule(lr){2-8}
                                     & \multirow{2}{*}{75\%}      & Vanilla CoT     & 62.14 & 64.02 & 26.94 & 50.69 & 50.95  \\
                                     &                             & Reverse         & 71.98 & \textbf{77.30} & 54.44 & 61.41 & 66.28  \\
                                     \arrayrulecolor{gray}\cmidrule(lr){2-8}
                                     & \multirow{2}{*}{100\%}     & Vanilla CoT     & 60.84 & 51.40 & 19.55 & 46.41 & 44.55  \\
                                     &                             & Reverse         & \textbf{72.05} & 76.96 & \textbf{59.21} & 59.17 & \textbf{66.85} \\
        \toprule[1.5pt]
    \end{tabular}}
    \vspace{-5mm}
    \label{tab:6}
\end{wrapfigure}

\noindent\textbf{Q5. What does performance on agentic and visual tasks indicate about teacher model suitability?} Table~\ref{tab:3}, which examines agentic and visual tasks, provides strong support for the small model learnability gap. This concept suggests that smaller student models (e.g., $\leq 3$B parameters) may not always benefit most from the largest available teachers, as they might learn more effectively from slightly smaller teachers whose reasoning complexity better matches their own capacity. Our results for the Qwen-2.5-VL-3B student on Visual-CoT clearly demonstrate this: distillation from smaller, capable teachers like GPT-4-mini ($45.44\%$ acc.) and o4-mini ($45.20\%$ acc.) leads to superior performance when compared to the largest GPT-4 ($42.92\%$ acc.). This implies that the CoTs from very large models like GPT-4 might be overly complex for a smaller, specialized model like Qwen-2.5-VL-3B to internalize effectively. The more digestible reasoning patterns of GPT-4-mini and o1-mini likely facilitate better knowledge transfer, highlighting that sheer teacher strength does not guarantee optimal distillation if the student struggles with the complexity. \textbf{Q6. Considering textual, agentic, and visual tasks, what general principles can be inferred for selecting an optimal teacher?} Several interesting observations lead to emerging principles: \textit{(1) The "Learnability Gap" Affects Smaller/Specialized Students.} For smaller or specialized students, the strongest teacher is not always the best. A teacher with more aligned reasoning complexity, even if smaller, can yield better results. \textit{(2) Student's Prior Distillation History Impacts Receptiveness.} The Llama-3.1-8B-R1 model, previously distilled from DeepSeek-R1, shows slightly lower average performance on textual tasks compared to base Llama-3.1-8B when further distilled by either Gemini-1.5-Pro or GPT-4. This suggests that a student's prior specializations or distillation experiences can hinder learning from new teachers if their strengths don't align, leading to less effective knowledge transfer.

\subsubsection{Student Architecture and Advanced Selection}

\begin{wrapfigure}{r}{0.45\textwidth}
    \vspace{-4mm} 
    \centering
    \small
    \captionof{table}{Performance on Visual Reasoning tasks across Dense (Qwen) and MoE (DeepSeek) architectures.}
    \vspace{-2mm}
    \resizebox{\linewidth}{!}{%
    \begin{tabular}{lccc}
        \toprule[1.2pt]
        Model & \multicolumn{2}{c}{Qwen-2.5 VL 8B (Dense)} & DeepSeek-VL2 (MoE) \\
        \cmidrule(lr){2-3} \cmidrule(lr){4-4}
        Dataset & OK-VQA & CLEVR & OK-VQA \\
        \hline 
        
        \multicolumn{4}{c}{\emph{\textbf{Data Augmentation}}} \\
        \midrule
        \rowcolor{orange!15}
        Zero Shot & 65.60 & 56.88 & 11.60 \\
        \rowcolor{orange!15}
        Zero Shot CoT & 61.84 & 58.36 & 12.92 \\ \rowcolor{orange!15}
        Vanilla CoT & 59.94 & 54.04 & 45.46 \\
        \hline
        
        \multicolumn{4}{c}{\emph{\textbf{Data Selection}}} \\
        \midrule  \rowcolor{purple!15}
        No Selection & 59.94 & 54.04 & 45.46 \\\rowcolor{purple!15}
        Teacher Filter & 63.80 & \textbf{67.60} & \textbf{51.82} \\\rowcolor{purple!15}
        Student Filter & \textbf{66.30} & 57.02 & 43.46 \\\rowcolor{purple!15}
        LLM Judge & 59.42 & 54.12 & 43.88 \\\rowcolor{purple!15}
        Model Uncertainty & 59.54 & 50.26 & 43.78 \\
        \hline
        
        \multicolumn{4}{c}{\emph{\textbf{Data Mixing}}} \\
         \rowcolor{cyan!15}
        \midrule
        No Mixing & 59.94 & 54.04 & 45.46 \\ \rowcolor{cyan!15}
        Teacher Mixing & 61.70 & 55.60 & 48.04 \\
        \bottomrule[1.2pt]
    \end{tabular}}
    \label{tab:moe_uncertainty}
    \vspace{-4mm} 
\end{wrapfigure}

While we discuss the scaling laws of standard dense student models in Appendix~\ref{app:student}, it is crucial to validate the universality of DC-CoT across diverse architectures and assess more complex data selection heuristics. To this end, we extended our evaluation to \textbf{DeepSeek-VL2} (a Mixture-of-Experts model) and \textbf{Qwen-2.5-VL-8B} on visual reasoning tasks (OK-VQA, CLEVR). Furthermore, we introduced an \textbf{Uncertainty-based Selection} strategy, which prioritizes training instances where the student model exhibits high entropy ($>0.5$) in zero-shot inference.

As presented in Table~\ref{tab:moe_uncertainty}, DC-CoT strategies remain effective for the MoE architecture. For instance, \textit{Teacher Filtering} improves DeepSeek-VL2's performance on OK-VQA from 45.46\% (Vanilla) to 51.82\%. Regarding data selection, while Uncertainty-based selection yields competitive results (e.g., 59.54\% on OK-VQA with Qwen), it does not consistently outperform our proposed heuristic methods (Student/Teacher Filtering). This suggests that the foundational primitives defined in DC-CoT are both robust and efficient for diverse student architectures including MoEs.


\subsection{Data-Level Results}

\subsubsection{Effect of Data Volume}
We investigate the relationship between the volume of seed data used for distillation and the resulting student model performance, referencing Table~\ref{tab:6} for the textual reasoning task with Gemini-1.5-Pro as the teacher and Reverse augmentation, and Figure~\ref{fig:bar} for agentic and visual tasks with Claude 3.5 as the teacher and CoT.

\noindent\textbf{Q7. How does increasing the percentage of seed data generally impact student model performance for Vanilla CoT and Reverse on textual tasks? How do these two methods compare?} On textual tasks, increasing seed data for Vanilla CoT does not consistently yield linear performance improvements. For Llama-3.1-8B, Vanilla CoT performance peaks at $50\%$ seed data, then declines. Mistral with Vanilla Cot shows a similar non-linear trend, peaking earlier at $25\%$. This suggests that additional raw teacher traces might introduce noise or less informative examples beyond an optimal point, potentially hindering learning. In contrast, Reverse augmentation generally shows more consistent benefits with increased data. For both models, Reverse results in better performance at higher data volumes. This indicates that the richer signal from Reverse is more effectively leveraged as data volume increases. \textbf{Q8. Does the ``more data always leads to better results" scaling law hold true across these experiments?} The traditional scaling law does not universally hold in our experiments. This is particularly evident for Vanilla CoT on textual tasks, where performance can degrade with excessive data. However, more data tends to be beneficial up to the tested volumes for more sophisticated augmentations like Reverse on textual data, and generally for agentic tasks.

\vspace{-4mm}
\subsubsection{Generalization Capability Analysis}
\vspace{-2mm}

\begin{wrapfigure}{r}{0.35\textwidth}
    \centering
    \small
    \vspace{-4mm}
    \captionof{table}{Zero-shot (ZS) versus OOD fine-tuning accuracy with Llama-3.1-8B.}
    \vspace{-2mm}
    \resizebox{\linewidth}{!}{%
    \begin{tabular}{lllc}
        \toprule[1.2pt]
        Training Data & Testing Data & Setting & ACC. \\
        \hline \addlinespace[2pt]
        \multirow{2}{*}{SQA} & \multirow{2}{*}{BoolQ} & ZS & 54.75 \\
         &  & OOD & \textbf{64.16} \\
        \hline \addlinespace[2pt]
        \multirow{2}{*}{ARC} & \multirow{2}{*}{OBQA} & ZS & 74.58 \\
         &  & OOD & \textbf{81.60} \\
        \hline \addlinespace[2pt]
        \multirow{2}{*}{ANLI} & \multirow{2}{*}{ESNLI} & ZS & 49.74 \\
         &  & OOD & \textbf{59.75} \\
        \hline \addlinespace[2pt]
        \multirow{4}{*}{GSM8K} & \multirow{2}{*}{GSM8K-Rev} & ZS & 16.74 \\
         &  & OOD & \textbf{38.89} \\
        \cmidrule(lr){2-4}
         & \multirow{2}{*}{MATH} & ZS & \textbf{9.32} \\
         &  & OOD & 8.75 \\
        \hline \addlinespace[2pt]
        \multirow{2}{*}{MATH} & \multirow{2}{*}{GSM8K} & ZS & 19.64 \\
         &  & OOD & \textbf{80.74} \\
        \hline \addlinespace[2pt]
        \multirow{2}{*}{Webarena-hard} & \multirow{2}{*}{Webarena-easy} & ZS & 14.18 \\
         &  & OOD & \textbf{19.90} \\
        \hline \addlinespace[2pt]
        \multirow{2}{*}{Webarena-easy} & \multirow{2}{*}{Webarena-hard} & ZS & 2.44 \\
         &  & OOD & \textbf{11.95} \\
        \hline \addlinespace[2pt]
        \multirow{2}{*}{Visual-CoT} & \multirow{2}{*}{OK-VQA} & ZS & \textbf{42.10} \\
         &  & OOD & 38.90 \\
        \hline \addlinespace[2pt]
        \multirow{2}{*}{OK-VQA} & \multirow{2}{*}{Visual-CoT} & ZS & 44.52 \\
         &  & OOD & \textbf{44.62} \\
        \hline \addlinespace[2pt]
        \toprule[1.2pt]
    \end{tabular}}
    \vspace{-7mm}
    \label{tab:4}
\end{wrapfigure}

We investigate how well reasoning skills learned through CoT distillation on a source dataset transfer to related but distinct target datasets. The analysis primarily references Table 4, while all experimental settings are explained in Appendix~\ref{app:gen}.

\noindent\textbf{9. How does fine-tuning on a source dataset generally impact Out-of-Distribution (OOD) performance compared to Zero-Shot performance on the target dataset?} Table~\ref{tab:4} consistently shows that fine-tuning on a source dataset, even if different from the target, generally leads to substantial improvements in OOD performance on the target dataset compared to its Zero-Shot accuracy. For instance, after training on SQA, OOD performance on BoolQ improves. Similarly, training on ARC boosts OBQA performance. This trend holds across textual, mathematical, agentic, and even some visual task pairings, indicating that the reasoning skills learned via CoT distillation possess a notable degree of transferability. \textbf{Q10. Are there specific task categories or pairings where OOD generalization is particularly strong or weak? Does fine-tuning on a source task always guarantee better OOD performance than its Zero-Shot counterpart on the target task?} The degree of generalization varies across task categories and specific pairings as observed in Table~\ref{tab:4}. Strong generalization is evident when transferring between similar textual reasoning tasks. For example, training on SQA significantly boosts BoolQ, and ARC training enhances OBQA performance. Mathematical reasoning also shows strong positive transfer, particularly when training on the more complex MATH dataset and testing on GSM8K, and also from GSM8K to its reversed version, GSM8K-Rev. Agentic tasks within WebArena also demonstrate good generalization across difficulty levels. However, generalization can be mixed or weak in other scenarios. For instance, while MATH to GSM8K is strong, the reverse (GSM8K to MATH) shows a decrease. Visual tasks also present varied results; training on OK-VQA improves Visual-Cot, but training on Visual-Cot leads to a drop on OK-VQA.

For a detailed analysis of the computational efficiency and token-level costs of our data-centric pipeline, please refer to Appendix~\ref{app:eff}.

\vspace{-4mm}
\section{Conclusion}
\vspace{-3mm}
This paper addresses the challenge of transferring reasoning from large to small models via CoT distillation, a domain where data-centric strategies have been underexplored. We introduce \textbf{DC-CoT}, a comprehensive benchmark designed to systematically investigate how data augmentation, selection, and mixing influence CoT distillation efficacy. Our findings reveal that data-centric manipulations significantly enhance distillation. Data augmentation, in particular, offers the most substantial performance gains by enriching the diversity of reasoning traces. Furthermore, we distill our findings into a heuristic framework for practitioners: (1) \textbf{Structured Logic tasks} (Math, Code) benefit most from \textbf{Reverse Thinking} combined with \textbf{Teacher Correctness} filtering to enforce logical consistency. (2) \textbf{Open-Ended Linguistic tasks} (Commonsense, NLI) require \textbf{Answer Augmentation} paired with \textbf{LLM-as-a-Judge} to capture diverse reasoning paths without semantic drift. (3) \textbf{Agentic and Visual tasks} necessitate \textbf{LLM-as-a-Judge} filtering, as simple heuristics fail to verify the grounding of rationales in observation contexts. Future work will expand this benchmark to include non-Transformer architectures and investigate more complex selection metrics, paving the way to democratize advanced reasoning.



\section*{Acknowledgement}
This work is generously supported by
Amazon Research Award, Cisco Faculty Award, UNC Accelerating AI Awards, NAIRR Pilot Award, OpenAI Researcher Access Award, and Gemma Academic Program GCP Credit Award.

\section*{Ethics Statement}
We adhere to the ICLR Code of Ethics. No private, sensitive, or personally identifiable data are involved. Our work does not raise foreseeable ethical concerns or produce harmful societal outcomes.

\section*{Reproducibility Statement}
Reproducibility is central to our work. All datasets used in our experiments are standard benchmarks that are publicly available. We provide full details of the training setup, model architectures, and evaluation metrics in the main paper and appendix. We have also released our codebase as anonymous repository, including scripts for preprocessing, training, and evaluation, along with configuration files and documentation to facilitate exact reproduction of our results. Random seeds and hyperparameters will also be included to further ensure reproducibility.

\bibliography{iclr2026_conference}

@article{wei2022chain,
  title={Chain-of-thought prompting elicits reasoning in large language models},
  author={Wei, Jason and Wang, Xuezhi and Schuurmans, Dale and Bosma, Maarten and Xia, Fei and Chi, Ed and Le, Quoc V and Zhou, Denny and others},
  journal={Advances in neural information processing systems},
  volume={35},
  pages={24824--24837},
  year={2022}
}

@inproceedings{zheng2024llamafactory,
  title={LlamaFactory: Unified Efficient Fine-Tuning of 100+ Language Models},
  author={Yaowei Zheng and Richong Zhang and Junhao Zhang and Yanhan Ye and Zheyan Luo and Zhangchi Feng and Yongqiang Ma},
  booktitle={Proceedings of the 62nd Annual Meeting of the Association for Computational Linguistics (Volume 3: System Demonstrations)},
  address={Bangkok, Thailand},
  publisher={Association for Computational Linguistics},
  year={2024},
  url={http://arxiv.org/abs/2403.13372}
}

@article{achiam2023gpt,
  title={Gpt-4 technical report},
  author={Achiam, Josh and Adler, Steven and Agarwal, Sandhini and Ahmad, Lama and Akkaya, Ilge and Aleman, Florencia Leoni and Almeida, Diogo and Altenschmidt, Janko and Altman, Sam and Anadkat, Shyamal and others},
  journal={arXiv preprint arXiv:2303.08774},
  year={2023}
}

@article{shao2024visual,
  title={Visual cot: Advancing multi-modal language models with a comprehensive dataset and benchmark for chain-of-thought reasoning},
  author={Shao, Hao and Qian, Shengju and Xiao, Han and Song, Guanglu and Zong, Zhuofan and Wang, Letian and Liu, Yu and Li, Hongsheng},
  journal={Advances in Neural Information Processing Systems},
  volume={37},
  pages={8612--8642},
  year={2024}
}

@article{zhou2023webarena,
  title={Webarena: A realistic web environment for building autonomous agents},
  author={Zhou, Shuyan and Xu, Frank F and Zhu, Hao and Zhou, Xuhui and Lo, Robert and Sridhar, Abishek and Cheng, Xianyi and Ou, Tianyue and Bisk, Yonatan and Fried, Daniel and others},
  journal={arXiv preprint arXiv:2307.13854},
  year={2023}
}

@inproceedings{marino2019ok,
  title={Ok-vqa: A visual question answering benchmark requiring external knowledge},
  author={Marino, Kenneth and Rastegari, Mohammad and Farhadi, Ali and Mottaghi, Roozbeh},
  booktitle={Proceedings of the IEEE/cvf conference on computer vision and pattern recognition},
  pages={3195--3204},
  year={2019}
}

@misc{o3o4mini2024,
  author       = {{OpenAI}},
  title        = {Introducing O3 and O4 Mini},
  year         = {2024},
  month        = may,
  url          = {https://openai.com/index/introducing-o3-and-o4-mini/},
  note         = {Accessed: 2025-05-15}
}

@misc{gpt4.1openai2024,
  author       = {{OpenAI}},
  title        = {GPT-4.1},
  year         = {2024},
  month        = may,
  url          = {https://openai.com/index/gpt-4-1/},
  note         = {Accessed: 2025-05-15}
}

@misc{claude3.5sonnet2024,
  author       = {{Anthropic}},
  title        = {Introducing Claude 3.5 Sonnet},
  year         = {2024},
  month        = jun,
  url          = {https://www.anthropic.com/news/claude-3-5-sonnet},
  note         = {Accessed: 2025-05-15}
}

@article{hinton2015distilling,
  title={Distilling the knowledge in a neural network},
  author={Hinton, Geoffrey and Vinyals, Oriol and Dean, Jeff},
  journal={arXiv preprint arXiv:1503.02531},
  year={2015}
}

@article{ho2022large,
  title={Large language models are reasoning teachers},
  author={Ho, Namgyu and Schmid, Laura and Yun, Se-Young},
  journal={arXiv preprint arXiv:2212.10071},
  year={2022}
}

@article{mukherjee2023orca,
  title={Orca: Progressive learning from complex explanation traces of gpt-4},
  author={Mukherjee, Subhabrata and Mitra, Arindam and Jawahar, Ganesh and Agarwal, Sahaj and Palangi, Hamid and Awadallah, Ahmed},
  journal={arXiv preprint arXiv:2306.02707},
  year={2023}
}

@article{chen2024reverse,
  title={Reverse Thinking Makes LLMs Stronger Reasoners},
  author={Chen, Justin Chih-Yao and Wang, Zifeng and Palangi, Hamid and Han, Rujun and Ebrahimi, Sayna and Le, Long and Perot, Vincent and Mishra, Swaroop and Bansal, Mohit and Lee, Chen-Yu and others},
  journal={arXiv preprint arXiv:2411.19865},
  year={2024}
}

@inproceedings{buciluǎ2006model,
  title={Model compression},
  author={Buciluǎ, Cristian and Caruana, Rich and Niculescu-Mizil, Alexandru},
  booktitle={Proceedings of the 12th ACM SIGKDD international conference on Knowledge discovery and data mining},
  pages={535--541},
  year={2006},
  organization={ACM}
}

@inproceedings{kim-etal-2023-aligning,
    title = "Aligning Large Language Models through Synthetic Feedback",
    author = "Kim, Sungdong  and
      Bae, Sanghwan  and
      Shin, Jamin  and
      Kang, Soyoung  and
      Kwak, Donghyun  and
      Yoo, Kang  and
      Seo, Minjoon",
    editor = "Bouamor, Houda  and
      Pino, Juan  and
      Bali, Kalika",
    booktitle = "Proceedings of the 2023 Conference on Empirical Methods in Natural Language Processing",
    month = dec,
    year = "2023",
    address = "Singapore",
    publisher = "Association for Computational Linguistics",
    url = "https://aclanthology.org/2023.emnlp-main.844/",
    doi = "10.18653/v1/2023.emnlp-main.844",
    pages = "13677--13700"
}

@article{tong2024optimizing,
  title={Optimizing Language Model's Reasoning Abilities with Weak Supervision},
  author={Tong, Yongqi and Wang, Sizhe and Li, Dawei and Wang, Yifan and Han, Simeng and Lin, Zi and Huang, Chengsong and Huang, Jiaxin and Shang, Jingbo},
  journal={arXiv preprint arXiv:2405.04086},
  year={2024}
}

@inproceedings{huang-etal-2023-large,
    title = "Large Language Models Can Self-Improve",
    author = "Huang, Jiaxin  and
      Gu, Shixiang  and
      Hou, Le  and
      Wu, Yuexin  and
      Wang, Xuezhi  and
      Yu, Hongkun  and
      Han, Jiawei",
    editor = "Bouamor, Houda  and
      Pino, Juan  and
      Bali, Kalika",
    booktitle = "Proceedings of the 2023 Conference on Empirical Methods in Natural Language Processing",
    month = dec,
    year = "2023",
    address = "Singapore",
    publisher = "Association for Computational Linguistics",
    url = "https://aclanthology.org/2023.emnlp-main.67/",
    doi = "10.18653/v1/2023.emnlp-main.67",
    pages = "1051--1068"
}

@inproceedings{wang-etal-2023-self-instruct,
    title = "Self-Instruct: Aligning Language Models with Self-Generated Instructions",
    author = "Wang, Yizhong  and
      Kordi, Yeganeh  and
      Mishra, Swaroop  and
      Liu, Alisa  and
      Smith, Noah A.  and
      Khashabi, Daniel  and
      Hajishirzi, Hannaneh",
    editor = "Rogers, Anna  and
      Boyd-Graber, Jordan  and
      Okazaki, Naoaki",
    booktitle = "Proceedings of the 61st Annual Meeting of the Association for Computational Linguistics (Volume 1: Long Papers)",
    month = jul,
    year = "2023",
    address = "Toronto, Canada",
    publisher = "Association for Computational Linguistics",
    url = "https://aclanthology.org/2023.acl-long.754/",
    doi = "10.18653/v1/2023.acl-long.754",
    pages = "13484--13508"
}

@inproceedings{wang-etal-2025-self,
    title = "Self-{DC}: When to Reason and When to Act? Self Divide-and-Conquer for Compositional Unknown Questions",
    author = "Wang, Hongru  and
      Xue, Boyang  and
      Zhou, Baohang  and
      Zhang, Tianhua  and
      Wang, Cunxiang  and
      Wang, Huimin  and
      Chen, Guanhua  and
      Wong, Kam-Fai",
    editor = "Chiruzzo, Luis  and
      Ritter, Alan  and
      Wang, Lu",
    booktitle = "Proceedings of the 2025 Conference of the Nations of the Americas Chapter of the Association for Computational Linguistics: Human Language Technologies (Volume 1: Long Papers)",
    month = apr,
    year = "2025",
    address = "Albuquerque, New Mexico",
    publisher = "Association for Computational Linguistics",
    url = "https://aclanthology.org/2025.naacl-long.331/",
    pages = "6510--6525",
    ISBN = "979-8-89176-189-6"
}

@article{lu2023self,
  title={Self: Self-evolution with language feedback},
  author={Lu, Jianqiao and Zhong, Wanjun and Huang, Wenyong and Wang, Yufei and Zhu, Qi and Mi, Fei and Wang, Baojun and Wang, Weichao and Zeng, Xingshan and Shang, Lifeng and others},
  journal={arXiv preprint arXiv:2310.00533},
  year={2023}
}

@misc{alpaca,
  author = {Rohan Taori and Ishaan Gulrajani and Tianyi Zhang and Yann Dubois and Xuechen Li and Carlos Guestrin and Percy Liang and Tatsunori B. Hashimoto },
  title = {Stanford Alpaca: An Instruction-following LLaMA model},
  year = {2023},
  publisher = {GitHub},
  journal = {GitHub repository},
  howpublished = {\url{https://github.com/tatsu-lab/stanford_alpaca}},
}

@article{xu2023wizardlm,
  title={Wizardlm: Empowering large language models to follow complex instructions},
  author={Xu, Can and Sun, Qingfeng and Zheng, Kai and Geng, Xiubo and Zhao, Pu and Feng, Jiazhan and Tao, Chongyang and Jiang, Daxin},
  journal={arXiv preprint arXiv:2304.12244},
  year={2023}
}

@article{zheng2023judging,
  title={Judging llm-as-a-judge with mt-bench and chatbot arena},
  author={Zheng, Lianmin and Chiang, Wei-Lin and Sheng, Ying and Zhuang, Siyuan and Wu, Zhanghao and Zhuang, Yonghao and Lin, Zi and Li, Zhuohan and Li, Dacheng and Xing, Eric and others},
  journal={Advances in Neural Information Processing Systems},
  volume={36},
  pages={46595--46623},
  year={2023}
}

@article{wang2022pinto,
  title={Pinto: Faithful language reasoning using prompt-generated rationales},
  author={Wang, Peifeng and Chan, Aaron and Ilievski, Filip and Chen, Muhao and Ren, Xiang},
  journal={arXiv preprint arXiv:2211.01562},
  year={2022}
}

@inproceedings{josifoski-etal-2023-exploiting,
    title = "Exploiting Asymmetry for Synthetic Training Data Generation: {S}ynth{IE} and the Case of Information Extraction",
    author = "Josifoski, Martin  and
      Sakota, Marija  and
      Peyrard, Maxime  and
      West, Robert",
    editor = "Bouamor, Houda  and
      Pino, Juan  and
      Bali, Kalika",
    booktitle = "Proceedings of the 2023 Conference on Empirical Methods in Natural Language Processing",
    month = dec,
    year = "2023",
    address = "Singapore",
    publisher = "Association for Computational Linguistics",
    url = "https://aclanthology.org/2023.emnlp-main.96/",
    doi = "10.18653/v1/2023.emnlp-main.96",
    pages = "1555--1574"
}

@article{zhang2023huatuogpt,
  title={Huatuogpt, towards taming language model to be a doctor},
  author={Zhang, Hongbo and Chen, Junying and Jiang, Feng and Yu, Fei and Chen, Zhihong and Li, Jianquan and Chen, Guiming and Wu, Xiangbo and Zhang, Zhiyi and Xiao, Qingying and others},
  journal={arXiv preprint arXiv:2305.15075},
  year={2023}
}

@article{zhao2023gimlet,
  title={Gimlet: A unified graph-text model for instruction-based molecule zero-shot learning},
  author={Zhao, Haiteng and Liu, Shengchao and Chang, Ma and Xu, Hannan and Fu, Jie and Deng, Zhihong and Kong, Lingpeng and Liu, Qi},
  journal={Advances in neural information processing systems},
  volume={36},
  pages={5850--5887},
  year={2023}
}

@article{ouyang2022training,
  title={Training language models to follow instructions with human feedback},
  author={Ouyang, Long and Wu, Jeffrey and Jiang, Xu and Almeida, Diogo and Wainwright, Carroll and Mishkin, Pamela and Zhang, Chong and Agarwal, Sandhini and Slama, Katarina and Ray, Alex and others},
  journal={Advances in neural information processing systems},
  volume={35},
  pages={27730--27744},
  year={2022}
}

@article{chen2020big,
  title={Big self-supervised models are strong semi-supervised learners},
  author={Chen, Ting and Kornblith, Simon and Swersky, Kevin and Norouzi, Mohammad and Hinton, Geoffrey E},
  journal={Advances in neural information processing systems},
  volume={33},
  pages={22243--22255},
  year={2020}
}

@article{kojima2022large,
  title={Large language models are zero-shot reasoners},
  author={Kojima, Takeshi and Gu, Shixiang Shane and Reid, Machel and Matsuo, Yutaka and Iwasawa, Yusuke},
  journal={Advances in neural information processing systems},
  volume={35},
  pages={22199--22213},
  year={2022}
}

@inproceedings{hsieh-etal-2023-distilling,
    title = "Distilling Step-by-Step! Outperforming Larger Language Models with Less Training Data and Smaller Model Sizes",
    author = "Hsieh, Cheng-Yu  and
      Li, Chun-Liang  and
      Yeh, Chih-kuan  and
      Nakhost, Hootan  and
      Fujii, Yasuhisa  and
      Ratner, Alex  and
      Krishna, Ranjay  and
      Lee, Chen-Yu  and
      Pfister, Tomas",
    editor = "Rogers, Anna  and
      Boyd-Graber, Jordan  and
      Okazaki, Naoaki",
    booktitle = "Findings of the Association for Computational Linguistics: ACL 2023",
    month = jul,
    year = "2023",
    address = "Toronto, Canada",
    publisher = "Association for Computational Linguistics",
    url = "https://aclanthology.org/2023.findings-acl.507/",
    doi = "10.18653/v1/2023.findings-acl.507",
    pages = "8003--8017"
}

@InProceedings{pmlr-v202-fu23d,
  title = 	 {Specializing Smaller Language Models towards Multi-Step Reasoning},
  author =       {Fu, Yao and Peng, Hao and Ou, Litu and Sabharwal, Ashish and Khot, Tushar},
  booktitle = 	 {Proceedings of the 40th International Conference on Machine Learning},
  pages = 	 {10421--10430},
  year = 	 {2023},
  editor = 	 {Krause, Andreas and Brunskill, Emma and Cho, Kyunghyun and Engelhardt, Barbara and Sabato, Sivan and Scarlett, Jonathan},
  volume = 	 {202},
  series = 	 {Proceedings of Machine Learning Research},
  month = 	 {23--29 Jul},
  publisher =    {PMLR},
  url = 	 {https://proceedings.mlr.press/v202/fu23d.html}
}

@inproceedings{li-etal-2023-symbolic,
    title = "Symbolic Chain-of-Thought Distillation: Small Models Can Also {\textquotedblleft}Think{\textquotedblright} Step-by-Step",
    author = "Li, Liunian Harold  and
      Hessel, Jack  and
      Yu, Youngjae  and
      Ren, Xiang  and
      Chang, Kai-Wei  and
      Choi, Yejin",
    editor = "Rogers, Anna  and
      Boyd-Graber, Jordan  and
      Okazaki, Naoaki",
    booktitle = "Proceedings of the 61st Annual Meeting of the Association for Computational Linguistics (Volume 1: Long Papers)",
    month = jul,
    year = "2023",
    address = "Toronto, Canada",
    publisher = "Association for Computational Linguistics",
    url = "https://aclanthology.org/2023.acl-long.150/",
    doi = "10.18653/v1/2023.acl-long.150",
    pages = "2665--2679"
}

@inproceedings{west-etal-2022-symbolic,
    title = "Symbolic Knowledge Distillation: from General Language Models to Commonsense Models",
    author = "West, Peter  and
      Bhagavatula, Chandra  and
      Hessel, Jack  and
      Hwang, Jena  and
      Jiang, Liwei  and
      Le Bras, Ronan  and
      Lu, Ximing  and
      Welleck, Sean  and
      Choi, Yejin",
    editor = "Carpuat, Marine  and
      de Marneffe, Marie-Catherine  and
      Meza Ruiz, Ivan Vladimir",
    booktitle = "Proceedings of the 2022 Conference of the North American Chapter of the Association for Computational Linguistics: Human Language Technologies",
    month = jul,
    year = "2022",
    address = "Seattle, United States",
    publisher = "Association for Computational Linguistics",
    url = "https://aclanthology.org/2022.naacl-main.341/",
    doi = "10.18653/v1/2022.naacl-main.341",
    pages = "4602--4625"
}

@inproceedings{magister-etal-2023-teaching,
    title = "Teaching Small Language Models to Reason",
    author = "Magister, Lucie Charlotte  and
      Mallinson, Jonathan  and
      Adamek, Jakub  and
      Malmi, Eric  and
      Severyn, Aliaksei",
    editor = "Rogers, Anna  and
      Boyd-Graber, Jordan  and
      Okazaki, Naoaki",
    booktitle = "Proceedings of the 61st Annual Meeting of the Association for Computational Linguistics (Volume 2: Short Papers)",
    month = jul,
    year = "2023",
    address = "Toronto, Canada",
    publisher = "Association for Computational Linguistics",
    url = "https://aclanthology.org/2023.acl-short.151/",
    doi = "10.18653/v1/2023.acl-short.151",
    pages = "1773--1781"
}

@article{mitra2023orca,
  title={Orca 2: Teaching small language models how to reason},
  author={Mitra, Arindam and Del Corro, Luciano and Mahajan, Shweti and Codas, Andres and Simoes, Clarisse and Agarwal, Sahaj and Chen, Xuxi and Razdaibiedina, Anastasia and Jones, Erik and Aggarwal, Kriti and others},
  journal={arXiv preprint arXiv:2311.11045},
  year={2023}
}

@inproceedings{li-etal-2023-making,
    title = "Making Language Models Better Reasoners with Step-Aware Verifier",
    author = "Li, Yifei  and
      Lin, Zeqi  and
      Zhang, Shizhuo  and
      Fu, Qiang  and
      Chen, Bei  and
      Lou, Jian-Guang  and
      Chen, Weizhu",
    editor = "Rogers, Anna  and
      Boyd-Graber, Jordan  and
      Okazaki, Naoaki",
    booktitle = "Proceedings of the 61st Annual Meeting of the Association for Computational Linguistics (Volume 1: Long Papers)",
    month = jul,
    year = "2023",
    address = "Toronto, Canada",
    publisher = "Association for Computational Linguistics",
    url = "https://aclanthology.org/2023.acl-long.291/",
    doi = "10.18653/v1/2023.acl-long.291",
    pages = "5315--5333"
}

@article{ding2024semcoder,
  title={Semcoder: Training code language models with comprehensive semantics reasoning},
  author={Ding, Yangruibo and Peng, Jinjun and Min, Marcus and Kaiser, Gail and Yang, Junfeng and Ray, Baishakhi},
  journal={Advances in Neural Information Processing Systems},
  volume={37},
  pages={60275--60308},
  year={2024}
}

@article{zelikman2022star,
  title={Star: Bootstrapping reasoning with reasoning},
  author={Zelikman, Eric and Wu, Yuhuai and Mu, Jesse and Goodman, Noah},
  journal={Advances in Neural Information Processing Systems},
  volume={35},
  pages={15476--15488},
  year={2022}
}

@article{lewkowycz2022solving,
  title={Solving quantitative reasoning problems with language models},
  author={Lewkowycz, Aitor and Andreassen, Anders and Dohan, David and Dyer, Ethan and Michalewski, Henryk and Ramasesh, Vinay and Slone, Ambrose and Anil, Cem and Schlag, Imanol and Gutman-Solo, Theo and others},
  journal={Advances in Neural Information Processing Systems},
  volume={35},
  pages={3843--3857},
  year={2022}
}

@article{yu2023metamath,
  title={Metamath: Bootstrap your own mathematical questions for large language models},
  author={Yu, Longhui and Jiang, Weisen and Shi, Han and Yu, Jincheng and Liu, Zhengying and Zhang, Yu and Kwok, James T and Li, Zhenguo and Weller, Adrian and Liu, Weiyang},
  journal={arXiv preprint arXiv:2309.12284},
  year={2023}
}

@article{li2024common,
  title={Common 7b language models already possess strong math capabilities},
  author={Li, Chen and Wang, Weiqi and Hu, Jingcheng and Wei, Yixuan and Zheng, Nanning and Hu, Han and Zhang, Zheng and Peng, Houwen},
  journal={arXiv preprint arXiv:2403.04706},
  year={2024}
}

@article{yuan2023scaling,
  title={Scaling relationship on learning mathematical reasoning with large language models},
  author={Yuan, Zheng and Yuan, Hongyi and Li, Chengpeng and Dong, Guanting and Lu, Keming and Tan, Chuanqi and Zhou, Chang and Zhou, Jingren},
  journal={arXiv preprint arXiv:2308.01825},
  year={2023}
}

@inproceedings{guo-etal-2024-exploring,
    title = "Exploring Reversal Mathematical Reasoning Ability for Large Language Models",
    author = "Guo, Pei  and
      You, WangJie  and
      Li, Juntao  and
      Bowen, Yan  and
      Zhang, Min",
    editor = "Ku, Lun-Wei  and
      Martins, Andre  and
      Srikumar, Vivek",
    booktitle = "Findings of the Association for Computational Linguistics: ACL 2024",
    month = aug,
    year = "2024",
    address = "Bangkok, Thailand",
    publisher = "Association for Computational Linguistics",
    url = "https://aclanthology.org/2024.findings-acl.811/",
    doi = "10.18653/v1/2024.findings-acl.811",
    pages = "13671--13685"
}

@inproceedings{10.1145/3097983.3098135,
author = {You, Shan and Xu, Chang and Xu, Chao and Tao, Dacheng},
title = {Learning from Multiple Teacher Networks},
year = {2017},
isbn = {9781450348874},
publisher = {Association for Computing Machinery},
address = {New York, NY, USA},
url = {https://doi.org/10.1145/3097983.3098135},
doi = {10.1145/3097983.3098135},
booktitle = {Proceedings of the 23rd ACM SIGKDD International Conference on Knowledge Discovery and Data Mining},
pages = {1285–1294},
numpages = {10},
location = {Halifax, NS, Canada},
series = {KDD '17}
}

@InProceedings{pmlr-v235-chen24ah,
  title = 	 {{MAGD}i: Structured Distillation of Multi-Agent Interaction Graphs Improves Reasoning in Smaller Language Models},
  author =       {Chen, Justin and Saha, Swarnadeep and Stengel-Eskin, Elias and Bansal, Mohit},
  booktitle = 	 {Proceedings of the 41st International Conference on Machine Learning},
  pages = 	 {7220--7235},
  year = 	 {2024},
  editor = 	 {Salakhutdinov, Ruslan and Kolter, Zico and Heller, Katherine and Weller, Adrian and Oliver, Nuria and Scarlett, Jonathan and Berkenkamp, Felix},
  volume = 	 {235},
  series = 	 {Proceedings of Machine Learning Research},
  month = 	 {21--27 Jul},
  publisher =    {PMLR},
  url = 	 {https://proceedings.mlr.press/v235/chen24ah.html}
}

@article{nye2021show,
  title={Show your work: Scratchpads for intermediate computation with language models},
  author={Nye, Maxwell and Andreassen, Anders Johan and Gur-Ari, Guy and Michalewski, Henryk and Austin, Jacob and Bieber, David and Dohan, David and Lewkowycz, Aitor and Bosma, Maarten and Luan, David and others},
  year={2021}
}

@article{joshi2023machine,
  title={Are machine rationales (not) useful to humans? measuring and improving human utility of free-text rationales},
  author={Joshi, Brihi and Liu, Ziyi and Ramnath, Sahana and Chan, Aaron and Tong, Zhewei and Nie, Shaoliang and Wang, Qifan and Choi, Yejin and Ren, Xiang},
  journal={arXiv preprint arXiv:2305.07095},
  year={2023}
}

@article{lanham2023measuring,
  title={Measuring faithfulness in chain-of-thought reasoning},
  author={Lanham, Tamera and Chen, Anna and Radhakrishnan, Ansh and Steiner, Benoit and Denison, Carson and Hernandez, Danny and Li, Dustin and Durmus, Esin and Hubinger, Evan and Kernion, Jackson and others},
  journal={arXiv preprint arXiv:2307.13702},
  year={2023}
}

@article{madaan2022text,
  title={Text and patterns: For effective chain of thought, it takes two to tango},
  author={Madaan, Aman and Yazdanbakhsh, Amir},
  journal={arXiv preprint arXiv:2209.07686},
  year={2022}
}

@inproceedings{wang-etal-2023-towards,
    title = "Towards Understanding Chain-of-Thought Prompting: An Empirical Study of What Matters",
    author = "Wang, Boshi  and
      Min, Sewon  and
      Deng, Xiang  and
      Shen, Jiaming  and
      Wu, You  and
      Zettlemoyer, Luke  and
      Sun, Huan",
    editor = "Rogers, Anna  and
      Boyd-Graber, Jordan  and
      Okazaki, Naoaki",
    booktitle = "Proceedings of the 61st Annual Meeting of the Association for Computational Linguistics (Volume 1: Long Papers)",
    month = jul,
    year = "2023",
    address = "Toronto, Canada",
    publisher = "Association for Computational Linguistics",
    url = "https://aclanthology.org/2023.acl-long.153/",
    doi = "10.18653/v1/2023.acl-long.153",
    pages = "2717--2739"
}

@article{dziri2023faith,
  title={Faith and fate: Limits of transformers on compositionality},
  author={Dziri, Nouha and Lu, Ximing and Sclar, Melanie and Li, Xiang Lorraine and Jiang, Liwei and Lin, Bill Yuchen and Welleck, Sean and West, Peter and Bhagavatula, Chandra and Le Bras, Ronan and others},
  journal={Advances in Neural Information Processing Systems},
  volume={36},
  pages={70293--70332},
  year={2023}
}

@article{yao2023tree,
  title={Tree of thoughts: Deliberate problem solving with large language models},
  author={Yao, Shunyu and Yu, Dian and Zhao, Jeffrey and Shafran, Izhak and Griffiths, Tom and Cao, Yuan and Narasimhan, Karthik},
  journal={Advances in neural information processing systems},
  volume={36},
  pages={11809--11822},
  year={2023}
}

@article{wang2022self,
  title={Self-consistency improves chain of thought reasoning in language models},
  author={Wang, Xuezhi and Wei, Jason and Schuurmans, Dale and Le, Quoc and Chi, Ed and Narang, Sharan and Chowdhery, Aakanksha and Zhou, Denny},
  journal={arXiv preprint arXiv:2203.11171},
  year={2022}
}

@article{madaan2023self,
  title={Self-refine: Iterative refinement with self-feedback},
  author={Madaan, Aman and Tandon, Niket and Gupta, Prakhar and Hallinan, Skyler and Gao, Luyu and Wiegreffe, Sarah and Alon, Uri and Dziri, Nouha and Prabhumoye, Shrimai and Yang, Yiming and others},
  journal={Advances in Neural Information Processing Systems},
  volume={36},
  pages={46534--46594},
  year={2023}
}

@inproceedings{yao2023react,
  title={React: Synergizing reasoning and acting in language models},
  author={Yao, Shunyu and Zhao, Jeffrey and Yu, Dian and Du, Nan and Shafran, Izhak and Narasimhan, Karthik and Cao, Yuan},
  booktitle={International Conference on Learning Representations (ICLR)},
  year={2023}
}

@article{shinn2023reflexion,
  title={Reflexion: Language agents with verbal reinforcement learning},
  author={Shinn, Noah and Cassano, Federico and Gopinath, Ashwin and Narasimhan, Karthik and Yao, Shunyu},
  journal={Advances in Neural Information Processing Systems},
  volume={36},
  pages={8634--8652},
  year={2023}
}

@article{liang2024sheep,
  title={I-SHEEP: Self-Alignment of LLM from Scratch through an Iterative Self-Enhancement Paradigm},
  author={Liang, Yiming and Zhang, Ge and Qu, Xingwei and Zheng, Tianyu and Guo, Jiawei and Du, Xinrun and Yang, Zhenzhu and Liu, Jiaheng and Lin, Chenghua and Ma, Lei and others},
  journal={arXiv preprint arXiv:2408.08072},
  year={2024}
}

@article{li2025small,
  title={Small models struggle to learn from strong reasoners},
  author={Li, Yuetai and Yue, Xiang and Xu, Zhangchen and Jiang, Fengqing and Niu, Luyao and Lin, Bill Yuchen and Ramasubramanian, Bhaskar and Poovendran, Radha},
  journal={arXiv preprint arXiv:2502.12143},
  year={2025}
}

@article{geva-etal-2021-aristotle,
    title = "Did Aristotle Use a Laptop? A Question Answering Benchmark with Implicit Reasoning Strategies",
    author = "Geva, Mor  and
      Khashabi, Daniel  and
      Segal, Elad  and
      Khot, Tushar  and
      Roth, Dan  and
      Berant, Jonathan",
    editor = "Roark, Brian  and
      Nenkova, Ani",
    journal = "Transactions of the Association for Computational Linguistics",
    volume = "9",
    year = "2021",
    address = "Cambridge, MA",
    publisher = "MIT Press",
    url = "https://aclanthology.org/2021.tacl-1.21/",
    doi = "10.1162/tacl_a_00370",
    pages = "346--361"
}

@inproceedings{talmor-etal-2019-commonsenseqa,
    title = "{C}ommonsense{QA}: A Question Answering Challenge Targeting Commonsense Knowledge",
    author = "Talmor, Alon  and
      Herzig, Jonathan  and
      Lourie, Nicholas  and
      Berant, Jonathan",
    editor = "Burstein, Jill  and
      Doran, Christy  and
      Solorio, Thamar",
    booktitle = "Proceedings of the 2019 Conference of the North {A}merican Chapter of the Association for Computational Linguistics: Human Language Technologies, Volume 1 (Long and Short Papers)",
    month = jun,
    year = "2019",
    address = "Minneapolis, Minnesota",
    publisher = "Association for Computational Linguistics",
    url = "https://aclanthology.org/N19-1421/",
    doi = "10.18653/v1/N19-1421",
    pages = "4149--4158"
}

@article{clark2018think,
  title={Think you have solved question answering? try arc, the ai2 reasoning challenge},
  author={Clark, Peter and Cowhey, Isaac and Etzioni, Oren and Khot, Tushar and Sabharwal, Ashish and Schoenick, Carissa and Tafjord, Oyvind},
  journal={arXiv preprint arXiv:1803.05457},
  year={2018}
}

@article{cobbe2021training,
  title={Training verifiers to solve math word problems},
  author={Cobbe, Karl and Kosaraju, Vineet and Bavarian, Mohammad and Chen, Mark and Jun, Heewoo and Kaiser, Lukasz and Plappert, Matthias and Tworek, Jerry and Hilton, Jacob and Nakano, Reiichiro and others},
  journal={arXiv preprint arXiv:2110.14168},
  year={2021}
}

@article{hendrycks2021measuring,
  title={Measuring mathematical problem solving with the math dataset},
  author={Hendrycks, Dan and Burns, Collin and Kadavath, Saurav and Arora, Akul and Basart, Steven and Tang, Eric and Song, Dawn and Steinhardt, Jacob},
  journal={arXiv preprint arXiv:2103.03874},
  year={2021}
}

@inproceedings{nie-etal-2020-adversarial,
    title = "Adversarial {NLI}: A New Benchmark for Natural Language Understanding",
    author = "Nie, Yixin  and
      Williams, Adina  and
      Dinan, Emily  and
      Bansal, Mohit  and
      Weston, Jason  and
      Kiela, Douwe",
    editor = "Jurafsky, Dan  and
      Chai, Joyce  and
      Schluter, Natalie  and
      Tetreault, Joel",
    booktitle = "Proceedings of the 58th Annual Meeting of the Association for Computational Linguistics",
    month = jul,
    year = "2020",
    address = "Online",
    publisher = "Association for Computational Linguistics",
    url = "https://aclanthology.org/2020.acl-main.441/",
    doi = "10.18653/v1/2020.acl-main.441",
    pages = "4885--4901"
}

@article{srivastava2022beyond,
  title={Beyond the imitation game: Quantifying and extrapolating the capabilities of language models},
  author={Srivastava, Aarohi and Rastogi, Abhinav and Rao, Abhishek and Shoeb, Abu Awal Md and Abid, Abubakar and Fisch, Adam and Brown, Adam R and Santoro, Adam and Gupta, Aditya and Garriga-Alonso, Adri{\`a} and others},
  journal={arXiv preprint arXiv:2206.04615},
  year={2022}
}

@article{team2024gemini,
  title={Gemini 1.5: Unlocking multimodal understanding across millions of tokens of context},
  author={Team, Gemini and Georgiev, Petko and Lei, Ving Ian and Burnell, Ryan and Bai, Libin and Gulati, Anmol and Tanzer, Garrett and Vincent, Damien and Pan, Zhufeng and Wang, Shibo and others},
  journal={arXiv preprint arXiv:2403.05530},
  year={2024}
}

@article{grattafiori2024llama,
  title={The llama 3 herd of models},
  author={Grattafiori, Aaron and Dubey, Abhimanyu and Jauhri, Abhinav and Pandey, Abhinav and Kadian, Abhishek and Al-Dahle, Ahmad and Letman, Aiesha and Mathur, Akhil and Schelten, Alan and Vaughan, Alex and others},
  journal={arXiv preprint arXiv:2407.21783},
  year={2024}
}

@article{guo2025deepseek,
  title={Deepseek-r1: Incentivizing reasoning capability in llms via reinforcement learning},
  author={Guo, Daya and Yang, Dejian and Zhang, Haowei and Song, Junxiao and Zhang, Ruoyu and Xu, Runxin and Zhu, Qihao and Ma, Shirong and Wang, Peiyi and Bi, Xiao and others},
  journal={arXiv preprint arXiv:2501.12948},
  year={2025}
}

@article{Jiang2023Mistral7,
  title={Mistral 7B},
  author={Albert Qiaochu Jiang and Alexandre Sablayrolles and Arthur Mensch and Chris Bamford and Devendra Singh Chaplot and Diego de Las Casas and Florian Bressand and Gianna Lengyel and Guillaume Lample and Lucile Saulnier and L'elio Renard Lavaud and Marie-Anne Lachaux and Pierre Stock and Teven Le Scao and Thibaut Lavril and Thomas Wang and Timoth{\'e}e Lacroix and William El Sayed},
  journal={ArXiv},
  year={2023},
  volume={abs/2310.06825},
  url={https://api.semanticscholar.org/CorpusID:263830494}
}

@article{team2024gemma,
  title={Gemma: Open models based on gemini research and technology},
  author={Team, Gemma and Mesnard, Thomas and Hardin, Cassidy and Dadashi, Robert and Bhupatiraju, Surya and Pathak, Shreya and Sifre, Laurent and Rivi{\`e}re, Morgane and Kale, Mihir Sanjay and Love, Juliette and others},
  journal={arXiv preprint arXiv:2403.08295},
  year={2024}
}

@article{yang2024qwen2,
  title={Qwen2. 5 technical report},
  author={Yang, An and Yang, Baosong and Zhang, Beichen and Hui, Binyuan and Zheng, Bo and Yu, Bowen and Li, Chengyuan and Liu, Dayiheng and Huang, Fei and Wei, Haoran and others},
  journal={arXiv preprint arXiv:2412.15115},
  year={2024}
}

@article{clark2019boolq,
  title={Boolq: Exploring the surprising difficulty of natural yes/no questions},
  author={Clark, Christopher and Lee, Kenton and Chang, Ming-Wei and Kwiatkowski, Tom and Collins, Michael and Toutanova, Kristina},
  journal={arXiv preprint arXiv:1905.10044},
  year={2019}
}

@inproceedings{OpenBookQA2018,
 title={Can a Suit of Armor Conduct Electricity? A New Dataset for Open Book Question Answering},
 author={Todor Mihaylov and Peter Clark and Tushar Khot and Ashish Sabharwal},
 booktitle={EMNLP},
 year={2018}
}

@article{camburu2018snli,
  title={e-snli: Natural language inference with natural language explanations},
  author={Camburu, Oana-Maria and Rockt{\"a}schel, Tim and Lukasiewicz, Thomas and Blunsom, Phil},
  journal={Advances in Neural Information Processing Systems},
  volume={31},
  year={2018}
}

@article{xu2024survey,
  title={A survey on knowledge distillation of large language models},
  author={Xu, Xiaohan and Li, Ming and Tao, Chongyang and Shen, Tao and Cheng, Reynold and Li, Jinyang and Xu, Can and Tao, Dacheng and Zhou, Tianyi},
  journal={arXiv preprint arXiv:2402.13116},
  year={2024}
}

@article{yang2025quantifying,
  title={Quantifying the Robustness of Retrieval-Augmented Language Models Against Spurious Features in Grounding Data},
  author={Yang, Shiping and Wu, Jie and Ding, Wenbiao and Wu, Ning and Liang, Shining and Gong, Ming and Zhang, Hengyuan and Zhang, Dongmei},
  journal={arXiv preprint arXiv:2503.05587},
  year={2025}
}

@inproceedings{tan2024large,
  title={Large Language Models for Data Annotation and Synthesis: A Survey},
  author={Tan, Zhen and Li, Dawei and Wang, Song and Beigi, Alimohammad and Jiang, Bohan and Bhattacharjee, Amrita and Karami, Mansooreh and Li, Jundong and Cheng, Lu and Liu, Huan},
  booktitle={Proceedings of the 2024 Conference on Empirical Methods in Natural Language Processing},
  pages={930--957},
  year={2024}
}

@article{li2024generation,
  title={From generation to judgment: Opportunities and challenges of llm-as-a-judge},
  author={Li, Dawei and Jiang, Bohan and Huang, Liangjie and Beigi, Alimohammad and Zhao, Chengshuai and Tan, Zhen and Bhattacharjee, Amrita and Jiang, Yuxuan and Chen, Canyu and Wu, Tianhao and others},
  journal={arXiv preprint arXiv:2411.16594},
  year={2024}
}

@article{li2025preference,
  title={Preference Leakage: A Contamination Problem in LLM-as-a-judge},
  author={Li, Dawei and Sun, Renliang and Huang, Yue and Zhong, Ming and Jiang, Bohan and Han, Jiawei and Zhang, Xiangliang and Wang, Wei and Liu, Huan},
  journal={arXiv preprint arXiv:2502.01534},
  year={2025}
}

@article{yu2025chain,
  title={Chain-of-Reasoning: Towards Unified Mathematical Reasoning in Large Language Models via a Multi-Paradigm Perspective},
  author={Yu, Yiyao and Zhang, Yuxiang and Zhang, Dongdong and Liang, Xiao and Zhang, Hengyuan and Zhang, Xingxing and Yang, Ziyi and Khademi, Mahmoud and Awadalla, Hany and Wang, Junjie and others},
  journal={arXiv preprint arXiv:2501.11110},
  year={2025}
}

@inproceedings{zhang2024balancing,
  title={Balancing speciality and versatility: a coarse to fine framework for supervised fine-tuning large language model},
  author={Zhang, Hengyuan and Wu, Yanru and Li, Dawei and Yang, Sak and Zhao, Rui and Jiang, Yong and Tan, Fei},
  booktitle={Findings of the Association for Computational Linguistics ACL 2024},
  pages={7467--7509},
  year={2024}
}

@inproceedings{johnson2017clevr,
  title={Clevr: A diagnostic dataset for compositional language and elementary visual reasoning},
  author={Johnson, Justin and Hariharan, Bharath and Van Der Maaten, Laurens and Fei-Fei, Li and Lawrence Zitnick, C and Girshick, Ross},
  booktitle={Proceedings of the IEEE conference on computer vision and pattern recognition},
  pages={2901--2910},
  year={2017}
}

@article{wang2024bpo,
  title={Bpo: Towards balanced preference optimization between knowledge breadth and depth in alignment},
  author={Wang, Sizhe and Tong, Yongqi and Zhang, Hengyuan and Li, Dawei and Zhang, Xin and Chen, Tianlong},
  journal={arXiv preprint arXiv:2411.10914},
  year={2024}
}

@article{li2025system,
  title={From system 1 to system 2: A survey of reasoning large language models},
  author={Li, Zhong-Zhi and Zhang, Duzhen and Zhang, Ming-Liang and Zhang, Jiaxin and Liu, Zengyan and Yao, Yuxuan and Xu, Haotian and Zheng, Junhao and Wang, Pei-Jie and Chen, Xiuyi and others},
  journal={arXiv preprint arXiv:2502.17419},
  year={2025}
}
\bibliographystyle{iclr2026_conference}

\appendix

\clearpage
\section{Limitations} \label{app:limit}
\begin{itemize}[leftmargin=*]
    \item \textbf{Budget Constraints:} Due to budget constraints, models like GPT-o4 were not included in our experiments. Moreover, migrating to other benchmarks also incurs substantial API costs. Therefore, for agentic task similar to many related papers~\citet{zhou2023webarena}, we focus solely on the \textsc{WebArena}~\citet{zhou2023webarena} framework. However, our method is simple and efficient, without any benchmark-specific optimizations, making it easily transferable to other models.

    \item \textbf{Hardware and Time Constraints:} Extending distillation to more and larger models is highly challenging due to hardware and time limitations. Therefore, we selected some student models for our distillation experiments.

\end{itemize}

\section{Broader Impact}
The DC-CoT benchmark is poised to significantly impact AI by fostering the development of smaller, more accessible, and powerful reasoning models. By systematically evaluating data-centric CoT distillation strategies, DC-CoT offers crucial insights and a standardized testbed, steering research towards resource-efficient AI and enabling advanced reasoning in computationally constrained environments. This research can yield broad societal and technological benefits:

\begin{enumerate}
    \item \textit{Democratization of AI:} Lowering computational barriers allows wider access to innovate with state-of-the-art AI.
    \item \textit{Educational Advancements:} Accessible reasoning models can be integrated into educational tools, supporting personalized learning.
    \item \textit{Application of AI:} Broader deployment of reasoning AI can aid complex problem-solving in research, healthcare, finance, and other industries.
\end{enumerate}

The insights from DC-CoT will also guide practitioners in optimizing distillation pipelines, promoting data-aware and sustainable AI by reducing the computational footprint of large models. By facilitating more efficient reasoning systems, DC-CoT contributes to a future of more equitably accessible and sustainably developed advanced AI.

\section{Experiment Setting} \label{app:exp}
\subsection{Distillation Training}
We conduct the distillation training on \textbf{8 A100 GPUs} and 16 \textbf{A6000 GPUs}, using LoRA fine-tuning for the student models. The LoRA rank we set is 32, and the lora alpha we set is 64. For an agentic task, the training process spans \textbf{5 epochs}, with a learning rate of $5*10^{-5}$ and a context length of $10,000$.  For the visual task, the training process spans \textbf{1 epoch}, with a learning rate of $5*10^{-5}$. The distillation methodology follows the guidelines provided in \texttt{Llama Factory}\citep{zheng2024llamafactory}. For Textual tasks, we train for 10 epochs for each dataset. 

\subsection{Inference Pipeline}
For inference, we employ the \texttt{vLLM} framework, running on \textbf{8 A100 GPUs}. The \textsc{WebArena} framework is deployed on \textbf{4 CPU machines}. To enhance efficiency, we leverage the official task-parallel Bash script for parallel execution, rather than processing tasks sequentially by task ID.

\subsection{Experimental Settings for Generalization Experiment}\label{app:gen}
For all experiments, we use Llama-3.1-8B as our student model. OOD Datasets were chosen as follows: BoolQ~\citep{clark2019boolq} was used for SQA, OBQA~\citep{OpenBookQA2018} for ARC, ESNLI~\citep{camburu2018snli} for ANLI, GSM8K-Rev~\citep{guo-etal-2024-exploring} and MATH for GSM8K, GSM8K for MATH, Webarena-easy for Webarena-hard and vice versa, Ok-VQA for Visual-CoT and vice versa. 

\section{Related Work} \label{app:related}

\subsection{Reasoning in LLMs}
\noindent The ability of LLMs to perform complex reasoning has been significantly enhanced by techniques that encourage explicit, step-by-step thinking. Foremost among these is Chain-of-Thought (CoT) prompting \citep{wei2022chain,kojima2022large, nye2021show}, which elicits intermediate reasoning steps from LLMs before arriving at a final answer. This approach makes the model's inference process more transparent by providing human-readable explanations \citep{joshi2023machine,lanham2023measuring} and substantially improves performance on tasks requiring multi-step deduction, such as arithmetic, commonsense, and symbolic reasoning \citep{wei2022chain}. By breaking down complex problems into manageable intermediate computations, CoT helps LLMs navigate intricate logical pathways and arrive at more accurate conclusions \citep{madaan2022text,wang-etal-2023-towards,dziri2023faith}. Integrating self-generated rationales through CoT effectively boosts the reasoning capabilities inherent in these models \citep{kojima2022large}. 

\noindent Building upon the foundational CoT paradigm, recent research has explored more sophisticated "deep-thinking" or "long-CoT" approaches to push the boundaries of LLM reasoning further. These methods often involve generating more elaborate or structured reasoning pathways. For example, Tree-of-Thought \citep{yao2023tree} prompting allows models to explore multiple reasoning paths in parallel, evaluating intermediate thoughts to decide the most promising direction. Other techniques focus on iterative refinement \citep{wang2022self} and self-correction, such as Self-Reflection \citep{madaan2023self,yao2023react,shinn2023reflexion}, where models critique and improve their own generated thoughts.

\subsection{Knowledge Distillation in LLMs}

\noindent Knowledge distillation is a potent technique for transferring knowledge from a large, often cumbersome, "teacher" model to a smaller, more efficient "student" model. This process is increasingly relevant in the context of LLMs due to their substantial size and computational demands. The fundamental concept, as introduced in early works \citep{buciluǎ2006model, hinton2015distilling}, involves training the student model to mimic the teacher model's output distribution (soft labels), thereby minimizing the divergence between their respective distributions. This approach has found applications across various tuning techniques for LLMs. For instance, LLM-generated annotations, including instructions, responses, and rationales, are leveraged in supervised fine-tuning, \textit{i.e.}, where a smaller model learns from the synthetic data produced by a larger teacher LLM \citep{kim-etal-2023-aligning,tong2024optimizing, huang-etal-2023-large,wang-etal-2023-self-instruct, wang-etal-2025-self, lu2023self}. This is particularly useful for enhancing specific capabilities \citep{josifoski-etal-2023-exploiting,zhang2023huatuogpt, zhao2023gimlet} or imparting domain-specific knowledge efficiently \citep{alpaca,xu2023wizardlm,zheng2023judging, wang2022pinto}. Furthermore, distillation techniques are employed in alignment tuning. One example includes Reinforcement Learning from Human Feedback (RLHF) \citep{ouyang2022training}, where synthetic data from LLMs can aid in reward modeling and policy training to align model outputs with human preferences and intentions. 

\noindent While classical knowledge distillation learns from the teacher model's distributions, and the objective is to minimize the difference between the student's distribution and the teacher's \citep{chen2020big}, recent advancements in LLMs have brought a particular focus to distilling their complex reasoning capabilities, especially CoT processes, into smaller student model's \citep{kojima2022large}. CoT is also crucial when addressing architectural differences or significant capacity gaps between teacher and student LLMs, as merely mimicking the final output might be insufficient for the student to learn effectively. Teacher models provide CoT rationales in various ways: (1) Sampled directly from the teacher \citep{hsieh-etal-2023-distilling,pmlr-v202-fu23d,li-etal-2023-symbolic,west-etal-2022-symbolic,magister-etal-2023-teaching,mukherjee2023orca,mitra2023orca}, (2) Generated via bootstrapping \citep{li-etal-2023-making,ding2024semcoder,zelikman2022star,lewkowycz2022solving,yu2023metamath,li2024common,yuan2023scaling,guo-etal-2024-exploring,chen2024reverse}, or (3) Obtained via multiple teacher models \citep{10.1145/3097983.3098135,pmlr-v235-chen24ah}. The rationale, reflecting the detailed thought process and reasoning pathway, serves as valuable auxiliary information for the student model to predict the final answer more accurately and robustly. 
While CoT distillation shows promise \citep{mukherjee2023orca,ho2022large}, it remains unclear which methods, teacher models are most effective for a specific student model and how they perform in various settings. This calls for a data-centric study of how the generation, selection, and combination of distillation data impact student reasoning and generalization.

\section{Task Descriptions} \label{app:tasks}

\noindent\textbf{Textual Reasoning:} It assesses a model's ability to make logical inferences from text, often through multi-step reasoning. Each instance includes a question $Q$, rationale $R$, and answer $A$. The student model $\mathcal{S}_\theta$ learns to predict $A$ using $Q$ and $R$. Tasks span commonsense, science, math, and table reasoning, with performance measured by answer accuracy.

\noindent\textbf{Agentic Reasoning:} This task tests an LLM agent $\pi_{\theta}$ in the \textsc{WebArena} browser sandbox, where it must follow an instruction $I$ by navigating real websites. At each step, the agent observes $o$, takes an action $a$, and explains its reasoning $r$. A large LLM ($M_{\mathrm{L}}$) selects actions based on the interaction history. Performance is measured by Success Rate (SR)—the fraction of tasks where the agent reaches the correct goal state.

\begin{wrapfigure}[13]{r}{0.5\textwidth} %
\vspace{-4mm}
    \centering
    \includegraphics[width=\linewidth]{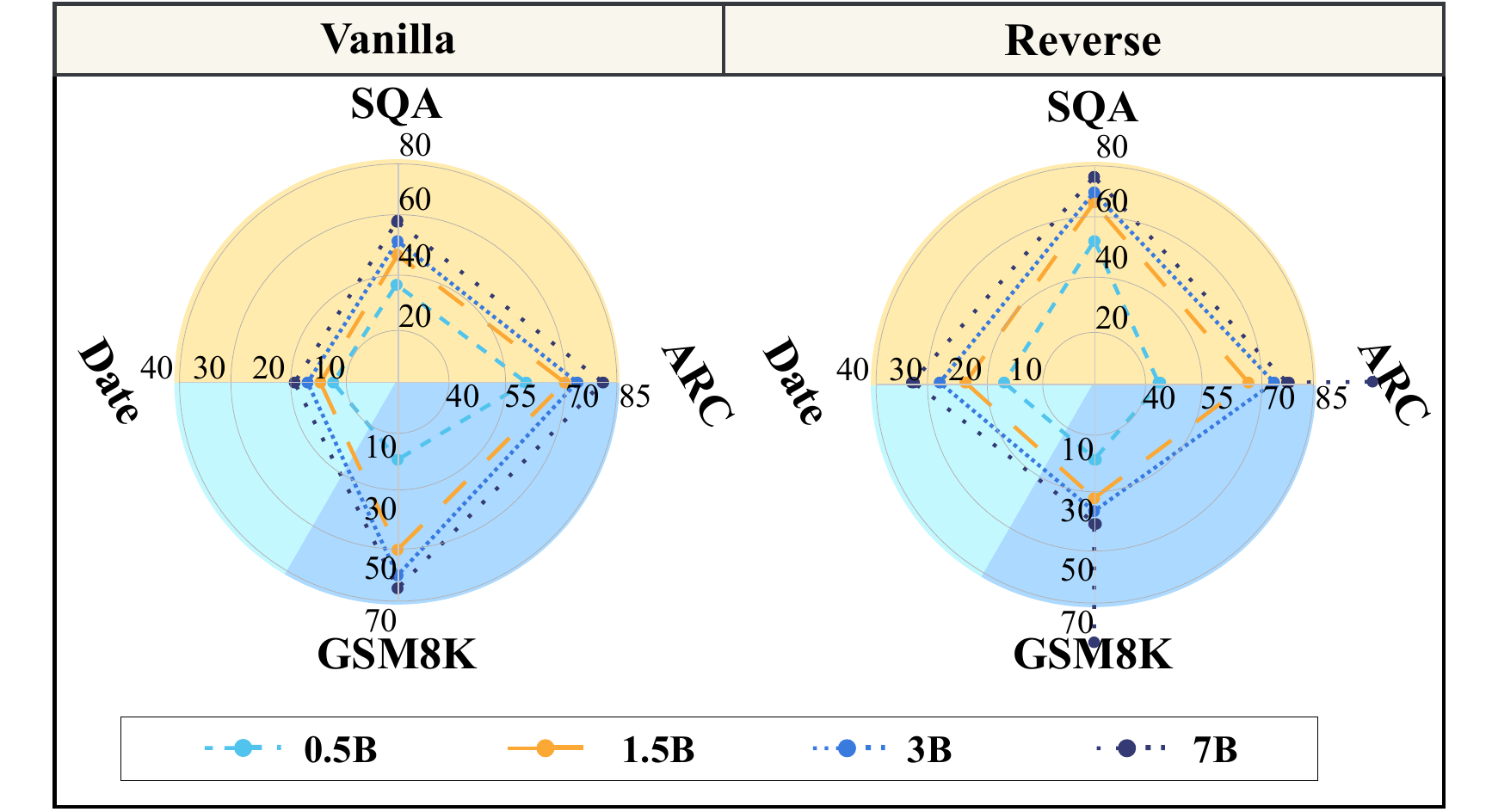} 
    \vspace{-3.5mm}
    \caption{Qwen-2.5 (0.5B-7B) distilled with Vanilla-CoT vs Reverse
    }
    \label{fig:radar} 
\end{wrapfigure}

\noindent\textbf{Visual Reasoning:} Extends chain-of-thought to multi-modal inputs, requiring models to interpret visual content and answer related questions. Each instance is a tuple $(v, q, a, r)$: an image $v$, a question $q$, an answer $a$, and a rationale $r$ outlining reasoning steps linking $v$ to $a$. Unlike text-only reasoning, visual reasoning demands interpretable grounding—$r$ often points to specific image regions that justify the answer. This keeps the reasoning process transparent, testing the model’s ability to connect visual cues with logical steps across multiple reasoning hops.

\section{Impact of Student Model Size} \label{app:student}
This section examines how the scale of the student model influences the efficacy of CoT distillation, with a particular focus on learnability from different augmentation strategies. The experiments, summarized in Figure~\ref{fig:radar}, are performed on Qwen-2.5 models of varying sizes (0.5B, 1.5B, 3B, 7B) when distilled with Vanilla CoT and Reverse augmentation, using Gemini-1.5-Pro as the teacher.

\noindent \textbf{Q11. How does the student model size generally affect reasoning performance with standard Vanilla CoT, and how does it interact with more complex augmentations like Reverse?} As shown in Figure~\ref{fig:radar}, performance with Vanilla CoT clearly scales with student model size: Qwen-2.5-0.5B achieves an average of $32.86\%$, which improves to $45.72\%$ for 1.5B, $50.89\%$ for 3B, and $55.58\%$ for the 7B model. This confirms that larger models better leverage standard teacher CoTs. The introduction of Reverse presents a more nuanced picture. On average across all four textual tasks, the impact is mixed; the 1.5B model shows a modest gain, while others see slight average decreases. However, these averages mask strong task-specific effects. Reverse significantly boosts performance on SQA and Date for all student sizes. Conversely, it markedly degraded performance on ARC and GSM8K compared to Vanilla CoT. This demonstrates that the utility of this complex augmentation is highly task-dependent in our specific student-teacher setup, instead of being a universal benefit. 

\noindent \textbf{Q12. Do smaller student models (0.5B, 1.5B) exhibit the small model learnability gap when faced with complex augmentations like Reverse?} The small model learnability gap suggests smaller models struggle with overly complex reasoning. Analyzing our results: On tasks where Reverse is beneficial, smaller models (0.5B, 1.5B) achieve substantial gains. However, their absolute scores remain below those of larger students, indicating a capacity limitation in reaching peak performance. 

\section{Efficiency and Code Analysis} \label{app:eff}
Efficiency is an important consideration for a data-centric pipeline designed for broad adoption. While wall-clock time can be a useful metric, it often varies significantly depending on the hardware, batch sizes, and API latencies. To offer a more hardware-independent and reproducible measure of computational cost, we instead report token-level costs for data generation. The token usage for several key prompting techniques is summarized in the Table~\ref{tab:token-costs}. 

Moreover, these costs are incurred once during data generation, with no inference-time overhead or change to the student model architecture. This design choice was deliberate: we aimed to make DC-CoT practical for both academic and applied ML use cases. 

\begin{table*}[h]
\centering
\caption{Token-level cost comparison for data generation methods.}
\label{tab:token-costs}
\resizebox{\textwidth}{!}{%
\begin{tabular}{l|l|c|c|c}
\toprule
\textbf{Method} & \textbf{Prompt Type} & \textbf{Avg. Prompt Tokens} & \textbf{Avg. Output Tokens} & \textbf{Total Tokens per Sample} \\
\midrule
Standard CoT & Forward CoT Prompt & 60 & 180 & 240 \\
Rephrased CoT & Question Rewriting & 75 & 180 & 255 \\
Reverse Thinking & Answer-First Reverse CoT & 110 & 200 & 310 \\
\bottomrule
\end{tabular}%
}
\end{table*}

To quantify the computational benefits of DC-CoT, we measured the throughput of our distilled students on a single NVIDIA A100-80GB GPU using vLLM. As shown in Table~\ref{tab:efficiency}, the distilled Qwen-2.5-3B model—which achieves performance competitive with larger vanilla baselines—offers a $\sim$6--9$\times$ speedup relative to the 8B baseline, validating the ``Efficient Reasoning'' claim of our benchmark.

\begin{table}[h]
\centering
\small
\caption{Efficiency profile of distilled student models measured on A100-80GB.}
\label{tab:efficiency}
\begin{tabular}{l c c c}
\toprule
\textbf{Model} & \textbf{VRAM (GB)} & \textbf{Throughput (tok/s)} & \textbf{Relative Speedup} \\
\midrule
Llama-3.1-8B (Student) & 16.2 & 115.4 & 1.0$\times$ \\
Qwen-2.5-3B (Student) & 7.8 & 184.2 & $\sim$1.6$\times$ \\
Qwen-2.5-1.5B (Student) & 4.2 & 245.1 & $\sim$2.1$\times$ \\
\bottomrule
\end{tabular}
\end{table}

\section{Confidence Intervals and Significance Testing}
While we reported average accuracy over 3 seeds for all experiments, we acknowledge that confidence intervals help contextualize gains that appear small. Some of these results are summarized in the Table~\ref{tab:confidence-intervals}. We report 95\% confidence intervals for core reasoning tasks (ARC, MATH, GSM8K) across Mistral-7B, LLaMA-3.1-8B, and Gemma-7B models. As shown below, Reverse CoT consistently and significantly outperforms No CoT, with non-overlapping intervals in nearly all cases, confirming the robustness of the gains. These results suggest that improvements are statistically significant, not noise.

\begin{table}[h]
\centering
\caption{Accuracy with 95\% confidence intervals on core reasoning tasks.}
\label{tab:confidence-intervals}
\small 
\begin{tabular}{l|l|p{0.20\columnwidth}|c}
\toprule
\textbf{Model} & \textbf{Task} & \textbf{Method} & \textbf{Accuracy ± CI} \\
\midrule
\multirow{6}{*}{Mistral-7B} & \multirow{2}{*}{ARC} & No CoT & 68.26 ± 0.75 \\
& & Reverse CoT & 76.96 ± 1.45 \\
\cmidrule{2-4}
& \multirow{2}{*}{MATH} & No CoT & 7.98 ± 0.39 \\
& & Reverse CoT & 16.12 ± 0.38 \\
\cmidrule{2-4}
& \multirow{2}{*}{GSM8K} & No CoT & 31.11 ± 1.80 \\
& & Reverse CoT & 59.21 ± 0.85 \\
\midrule
\multirow{6}{*}{LLaMA-3.1-8B} & \multirow{2}{*}{ARC} & No CoT & 60.41 ± 1.37 \\
& & Reverse CoT & 82.17 ± 1.20 \\
\cmidrule{2-4}
& \multirow{2}{*}{MATH} & No CoT & 7.39 ± 0.13 \\
& & Reverse CoT & 35.52 ± 0.26 \\
\cmidrule{2-4}
& \multirow{2}{*}{GSM8K} & No CoT & 20.74 ± 1.04 \\
& & Reverse CoT & 76.35 ± 1.98 \\
\midrule
\multirow{6}{*}{Gemma-7B} & \multirow{2}{*}{ARC} & No CoT & 68.09 ± 1.17 \\
& & Reverse CoT & 73.46 ± 1.06 \\
\cmidrule{2-4}
& \multirow{2}{*}{MATH} & No CoT & 7.24 ± 0.34 \\
& & Reverse CoT & 16.54 ± 0.83 \\
\cmidrule{2-4}
& \multirow{2}{*}{GSM8K} & No CoT & 26.22 ± 0.83 \\
& & Reverse CoT & 53.45 ± 1.75 \\
\bottomrule
\end{tabular}
\end{table}



\clearpage
\section{Thinking Example} \label{app:think}

\begin{tcolorbox}[breakable]
\small
\ttfamily
\lstset{
    basicstyle=\ttfamily\small, 
    breaklines=true, 
    breakatwhitespace=false, 
    columns=fullflexible, 
    backgroundcolor=\color{gray!10}, 
    keywordstyle=\color{blue}, 
    escapeinside={(*@}{@*)}, 
}

\textbf{Instruction :}
\begin{lstlisting}
        You need to issue an action,interaction history summary for this step. When you thinking need have OBSERVATION DESCRIPTION, OBSERVATION HIGHLIGHT, REASON.
        
        You are ONLY allowed to use the following action commands. Strictly adheres to the given format. 
        
        {click [id]: To click on an element with its numerical ID on the webpage. E.g., `click [7]`  ;
        
        type [id] [content] [press_enter_after=0|1]: To type content into a field with a specific ID. By default, the \"Enter\" key is pressed after typing unless `press_enter_after` is set to 0. E.g., `type [15] [Carnegie Mellon University] [1]` .;
        
        stop [answer]: To stop interaction and return response. Present your answer within the brackets. If the task doesn't require a textual answer or appears insurmountable, additional reasons and all relevant information you gather as the answer. E.g., `stop [N/A ...]`. ;
        
        note [content]: To take note of all important info w.r.t. completing the task to enable reviewing it later. E.g., `note [Spent $10 on 4/1/2024]`;
        
        go_back: To return to the previously viewed page.
\end{lstlisting}

\textbf{Input :}
\begin{lstlisting}
        OBJECTIVE: Which US states border Vermont?
        
        LASTACTION: first step, no last action
        
        LAST INTERACTION HISTORY SUMMARY: This is the initial interaction. We are on the OpenStreetMap homepage and need to search for Vermont to begin our task of identifying its bordering states.
        
        CURRENT OBSERVATION: 
            RootWebArea "OpenStreetMap"
            link [41]  'OpenStreetMap logo OpenStreetMap'
            link [402] 'Edit'
            button [403]
            link [373] 'History'
            link [374] 'Export'
            link [407] 'GPS Traces'
            link [408] 'User Diaries'
            link [409] 'Communities'
            link [410] 'Copyright'
            link [411] 'Help'
            link [412] 'About'
            link [382] 'Log In'
            link [383] 'Sign Up'
            link [515] 'Where is this?'
            textbox [35] 'Search' [required: False]
            button [516] 'Go'
            link [503] 'Find directions between two points'
            heading      'Welcome to OpenStreetMap!'
            button [469] 'Close'
            text         'OpenStreetMap is a map of the world, created by people like you and free to use under an open license.'
            text         'Hosting is supported by'
            link [475]  'UCL'
            link [477]  'Fastly'
            link [479]  'Bytemark Hosting'
            text         ', and other'
            link [481]  'partners'
            link [485]  'Learn More'
            link [486]  'Start Mapping'
            generic [15] 'Zoom In Zoom Out Show My Location Layers Share 50 km 50 mi OpenStreetMap contributors Make a Donation. Website and API terms'
                link [19]  'Zoom In'
                link [20]  'Zoom Out'
                button [22] 'Show My Location'
                link [24]  'Layers'
                link [296]
                link [26]  'Share'
                link [309] 'OpenStreetMap contributors'
                link [311] 'Make a Donation'
                link [313] 'Website and API terms'

\end{lstlisting}

\textbf{Output :}
\begin{lstlisting}

        "<think>
            OBSERVATION DESCRIPTION: The current page is the OpenStreetMap homepage. It contains a search box (ID: 35) where we can enter our query. There are various navigation links and buttons, but the most relevant element for our task is the search functionality. The map view (ID: 15) is also present, which will be crucial for visualizing Vermont and its neighboring states once we perform the search.
            
            OBSERVATION HIGHLIGHT: 35, 15, 516, 19, 20, 24, 26
            
            REASON: To locate Vermont on the map, we need to use the search functionality.
        
        </think>
        ACTION: type [35] [Vermont] [1]
        
        INTERACTION HISTORY SUMMARY: We have searched for \"Vermont\" on OpenStreetMap and are now looking at the search results page with a map view of Vermont."

\end{lstlisting}
\end{tcolorbox}

\clearpage
\section{Dataset Statistics}
\begin{table}[h!]
    \small
    \centering
    \begin{tabular}{lccccc}
    \toprule
         Dataset & Domain & License & Train (Original) & Train (Filtered) & Test \\\midrule
         SQA \cite{geva-etal-2021-aristotle} & Commonsense & MIT & 2,061 & 1,544 & 229 \\
         CSQA \cite{talmor-etal-2019-commonsenseqa}& Commonsense & MIT & 9,741 & 6,478 & 1,140 \\
         ARC \cite{clark2018think} & Commonsense & CC BY-SA 4.0 & 1,199 & 1,035 & 1,172 \\
         BoolQ \cite{clark2019boolq} & Commonsense & CC BY-SA 3.0 & 9,427 & 0 & 3,270 \\
         OpenbookQA \cite{OpenBookQA2018} & Commonsense & Apache & 4957 & 0 & 500 \\
         MATH \cite{hendrycks2021measuring} & Math & MIT & 7,500 & 2,511 & 5,000 \\
         GSM8K \cite{cobbe2021training} & Math & MIT & 7,379 & 4,293 & 1,339 \\
        GSM8K-Rev \cite{guo-etal-2024-exploring} & Math & Apache & - & 0 & 777 \\
         ANLI (r3) \cite{nie-etal-2020-adversarial} & NLI & CC BY-NC 4.0 & 100,459 & 883 & 1,200 \\
        e-SNLI \cite{camburu2018snli} & NLI & CC BY-NC 4.0 & 549,367 & 0 & 9,824 \\
         Date \cite{srivastava2022beyond} & Logic & Apache & - & 200 & 169 \\
         Webarena \cite{zhou2023webarena} & Agentic & Apache & - & 0 & 812 \\
         Visual-CoT \cite{shao2024visual} & Visual & Apache & 132,000 & 943,000 & 12,500 \\
         OK-VQA \cite{marino2019ok} & Visual & CC BY 4.0 & 5,046 & 9,009 & 5,000 \\
         \bottomrule
    \end{tabular}
    \caption{The datasets used in our Experimental Setup.}
    \label{tab:statistics}
\end{table}

\section{The Use of Large Language Models (LLMs)}
To enhance clarity and readability, we employed OpenAI's GPT-5 and GPT-5-thinking models exclusively as language polishing tools. Their role was limited to proofreading, grammatical correction, and stylistic refinement—functions comparable to those of conventional grammar checkers and dictionaries. These tools did not contribute any new scientific content or ideas, and their usage is consistent with standard practices in manuscript preparation.

\end{document}